\documentclass{article}
\usepackage[english]{babel}

\usepackage[letterpaper,top=2cm,bottom=2cm,left=3cm,right=3cm,marginparwidth=1.75cm]{geometry}
\usepackage{indentfirst}
\usepackage{makecell}
\usepackage{longtable}
\usepackage{amsmath}
\usepackage{mathptmx}
\usepackage{graphicx}
\usepackage{booktabs}
\usepackage{amssymb}
\usepackage{float}
\usepackage{verbatim}
\usepackage{xcolor}
\usepackage[export]{adjustbox}
\usepackage[linesnumbered,ruled,vlined]{algorithm2e}
\usepackage{multirow}
\usepackage[colorlinks=true, allcolors=blue]{hyperref}  
\usepackage{array}
\usepackage{subcaption}
\newcommand{\Rmnum}[1]{\expandafter\@slowromancap\romannumeral #1@}
\makeatother
\usepackage{CJK}
\usepackage{colortbl}
\usepackage{longtable}
\usepackage[T1]{fontenc}
\usepackage[utf8]{inputenc}
\usepackage{authblk}

\usepackage{amssymb}

\usepackage[colorlinks=true, allcolors=blue]{hyperref}
\title{RepAct: The Re-parameterizable Adaptive Activation Function}
\author[1]{Xian Wu}
\author[1]{Qingchuan Tao\thanks{Corresponding author: 451911031@qq.com}}
\author[1]{Shuang Wang}

\affil[1]{College of Electronics and Information Engineering, Sichuan University, Chengdu, China}

\date{}

\begin{document}
\begin{sloppypar}
\maketitle

\begin{abstract}
With the ever-increasing demand and application of Internet of Things (IoT) technology in the real world, artificial intelligence applications in edge computing have attracted more and more attention. However, due to the limitations in computational resources on edge devices, the design of neural networks at the edge often must consider aspects such as lightweight structure and logical reasoning capabilities. While there has been considerable research on lightweight neural network architectures for the edge, little attention has been paid to how to enhance the reasoning capabilities of neural networks within the constraints of limited model parameters. To address this issue, we propose RepAct (Re-parameterizable Adaptive Activation Function), a simple yet effective adaptive activation function that can be re-parameterized, aimed at fully utilizing model parameter capacity while enhancing the inference and understanding abilities of lightweight networks. Specifically, RepAct adopts a multi-branch activation function structure to exploit different features information through various activation branches. Furthermore, to improve the understanding for the different activation branches, RepAct integrates each branch across different layers of the neural network with learnable adaptive weights. RepAct, by training a variety of power-function-based activation functions like HardSwish and ReLU in multi-branch settings, has been validated on a range of tasks including image classification, object detection, and semantic segmentation. It demonstrated significant improvements over the original lightweight network activation functions, including a 7.92\% Top-1 accuracy increase on MobileNetV3-Small on ImageNet100. At the same time, with computational complexity at inference stage approximating HardSwish, RepAct approached or even surpassed the task accuracy achieved by various mainstream activation functions and their variants in datasets like Cifar100, Cifar10, VOC12-Detect and VOC12-Segment within MobileNetV3, proving the effectiveness and advanced nature of the proposed method.

\end{abstract}


\section{Introduction}

Since the inception of the neural network resurgence with AlexNet\cite{1}, the design of activation functions has garnered continuous attention \cite{2}. Activation functions introduce non-linearity into the networks and play a critical role in feature extraction. In traditional neural network designs, the selection and design of activation functions for different layers and networks are typically based on manual design experience \cite{3,4} or adapted through NAS (Neural Architecture Search) \cite{5,6}. The emergence of adaptive activation function \cite{6,7,8} effectively improves network performance.   Subsequently, various dynamic adaptive parameters were introduced from different dimensions of the feature graph \cite{9,10,11}. Although the introduced parameters and computation amount were small, the memory cost of element by element operation often formed a bottleneck of lightweight network reasoning \cite{12}.

When deploying lightweight networks on resource-constrained edge devices, real-time performance requirements impose strict limitations on model parameters, computational power, and memory operations \cite{12,13,14,15}. Convolutional neural networks exhibit sparsity in their activations \cite{16}, which prevents lightweight networks from fully utilizing model capacity to learn features from task data. We have noticed that re-parameterizable convolutional structures \cite{17,18,19,20,21}, by virtue of their multi-branch architecture during training, enhance the network's feature capturing ability. At inference stage, these networks leverage the linear computational characteristics to revert to a single-branch model for deployment, balancing network performance with inference speed, thereby rejuvenating various classical network architectures.
Inspired by this, we propose a series of re-parameterizable adaptive activation functions, RepAct, which, through multi-branch activation during training and single-branch activation during inference, enhance the learning capabilities of lightweight networks in various tasks without additional memory access and computational overhead. RepAct adaptively adjusts the weights of each branch within different layers of the network according to gradient descent. We adopted common activation functions from lightweight networks to form a multi-branch activation structure, thus enhancing the feature extraction capability of the current network layer. Also, we introduce an Identity linear branch that avoids the destruction of feature information caused by nonlinear activations, balancing the linear and nonlinear flow of feature information, making it easier for features to transmit across network layers. This was analyzed and verified from the perspective of inter-layer feature propagation and gradient scaling. In addition, we have further explored RepAct-II with a degradable soft gating mechanism (Softmax-type) and RepAct-III with global degradable information (BN-type) to adapt to different types of task scenarios.

We substituted the original activation functions in the backbone of lightweight networks (ShuffleNet, MobileNet, and lightweight ViT) with the RepAct series of activation functions and validated their performance on image classification, object detection, and segmentation tasks against the original networks. After substituting with the RepAct series, a considerable improvement in task accuracy was observed compared to the original and other adaptive activation functions, which verified the capability enhancement of the RepAct series for feature extraction in lightweight networks. Moreover, by utilizing GradCAM and RepAct visualization, we analyzed the reasons for the improvement in network feature extraction ability from forward inference and gradient back-propagation.

This paper addresses the performance degradation issue in lightweight networks by first proposing a plug-and-play, re-parameterizable adaptive activation function that significantly improves the learning abilities of various lightweight networks in their respective task training without almost any increase at the inference stage. Secondly, we validated and analyzed the characteristics of RepAct in feature transmission and gradient solution, deeply observing and elucidating its mechanism of action, providing a theoretical basis for network optimization. We designed RepAct-II and RepAct-III with a degradable soft gating mechanism and degradable global information, respectively, suitable for different tasks and network scenarios, hence enhancing the model's flexibility and universality.

\section{RELATED WORKS}

\subsection{Activation function}

Different types of activation functions significantly affect the feature extraction capabilities and gradient propagation of neural networks, such as ReLU \cite{1}, SoftPlus \cite{22}, ELU \cite{23}, Mish \cite{24}, and their variants \cite{8,9,25,26,27}. These activation functions often focus on discussions about continuity, differentiability, and the saturation of the upper and lower bounds. In the design of network structures, we have a substantial and excellent library of activation functions at our disposal. However, the design of activation functions for different network structures in various datasets still relies on manual experience \cite{3,4} or NAS searches \cite{5,6}. Each attempt comes with considerable cost; thus, how to design an activation function suitable for the current data distribution and network structure remains a long-standing problem in network design.

MobileNetV2 \cite{3} discusses the harms of overly strong activation functions when the feature map channel dimensions are narrow in an inverted residual structure, confirming that the realms of activation layer and network architectural design complement each other. VallinaNet \cite{11} greatly enhances the feature extraction ability of shallow networks by introducing additional stacking calculations of activation functions. Networks composed solely of ReLU can be equivalent to a 3-layer shallow network during inference \cite{28}, but it is still worth pondering which activations should be chosen during training to optimize the feature extraction and flow between different network layers, enabling the network to reach the optimum via gradient descent.

Learnable activation functions can be involved in backward propagation as network parameters, endowing the network with a stronger learning capacity. PReLU \cite{7}, as an adaptive improvement of LReLU, introduces a learnable slope parameter to participate in training and achieves human-level performance on the ImageNet dataset \cite{29} with virtually no additional inference cost. FReLU \cite{10} uses a 2D funnel-like spatial gating mechanism to enable dynamic spatial modeling activation capability. DYReLU \cite{9} activates after dynamically calculating the current feature map through a module similar to SE \cite{30}, which is particularly effective for lightweight networks. AReLU \cite{8} acknowledges the gradient amplification of the activation layer as a vital characteristic for the rapid convergence of neural networks. ACON \cite{31}, by adaptively choosing active neurons (dynamically), performs well in terms of precision on lightweight networks and presents a series of designs with different granularity and dynamic-static combinations. Sheng \cite{32} explored gated, mixed, and hierarchical combinations of linear and exponential activation functions and designed combinatory adaptive activation functions. Although most dynamic adaptive activation functions effectively improve the task accuracy of networks, compensate for the insufficient model capacity and weak learning ability of lightweight networks, the computational complexity of the exponential type is too high, and dynamic adaptiveness relies on additional memory operations. Access to the feature map causes redundant serial memory overhead due to traversal or fragmented access, hindering the fusion of computational modules and activation operators, disastrously impacting inference speed on edge devices \cite{12}.

\subsection{Reparameterable structure}

Re-parameterizable neural network structures \cite{17,18,20,33,34} and re-parameterizable convolutional structures separate the multi-branch network architecture during training from the single-branch during deployment. They aim to capture as much information as possible during the training phase, while capitalizing on linear characteristics to merge branches during the deployment phase. Apple's MobileOne \cite{19} re-parameterizes the MobileNet architecture to achieve a backbone network with a phone inference duration at the millisecond level. RepViT \cite{21} undertakes re-parameterization from a ViT perspective, surpassing existing state-of-the-art lightweight ViTs. RepOptimizer \cite{35} applies gradient re-parameterization to the VGG \cite{36} style models by modifying gradients based on specific hyperparameters, incorporating model-specific prior knowledge into the optimizer, focusing on effective training. However, to date, the design of re-architecturable structures has concentrated on dense linear computational modules and has not fully explored the re-parameterizable form of activation functions, which constitute an essential part of the non-linear operations in a network model. Thus, exploring the re-parameterizable form of activation functions is crucial for enhancing the capability of networks to extract features.

\section{RepAct}
The RepAct proposed in this article is a re-parameterizable adaptive activation function. It utilizes a combination of common activation functions for multi-branch training during the training stage, where each branch's weight factor is adaptively adjusted through gradient descent. In the inference stage, these branch weight factors are re-parameterized in different segments to revert to a single-branch structure. The paper analyzes how the adaptively scaled adjustments of RepAct I to the forward features and the backward gradients enhance the network's learning capability. Furthermore, it proposes RepAct III-Softmax with a degradable soft gating mechanism, and RepAct III-BN with degradable global information.

\subsection{RepAct}

\begin{figure}[h]
\centering
\includegraphics[width=1\linewidth]{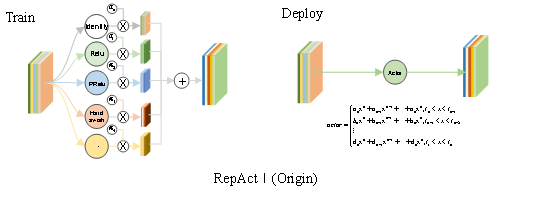}
\caption{\label{fig:frog}Schematic diagram of the single-branch structure of RepAct I} training multi-branch inference.
\end{figure}
The RepAct I structure is a universal type that combines common activation functions into a multi-branch setup, summing computed feature maps across each branch. This achieves a fusion of feature information mapped after activation under different nonlinear branches, thereby enhancing the network's feature extraction capability.

The RepAct I structure presents two distinct paradigms: a multi-branch structure during training and a re-parameterized single-branch structure during deployment. When the selected branch activation functions are all of the power-function type, the degree of re-parameterization for each branch segment is higher. Hence, this paper opts for power-function type activation functions that are suitable for deployment in lightweight networks. In terms of activation function segmentation, the empirical selection inherits the union of segments from each branch. After re-parameterization to a single branch, the memory access consumption during inference remains the same as for non-dynamic calculation class activation functions, equating to the size of the input feature map. The computational complexity is approximated to the complexity required for the RepAct's largest branch, as detailed in equation 1.
\begin{equation}
   O(RepAct) \approx max(O(RepAct_n)) 
\end{equation}

O(RepAct\_n) Is the computational complexity of the N\^{th} brahch of RepAct

In some network architectures, nonlinear operations can disrupt the transmission of feature information between layers. MobileNetV2 \cite{3} mentions that in certain layers, the nonlinear behavior of activation functions may damage the feature structure, resulting in a loss of feature information and a decrease in network performance. Therefore, in the RepAct structure, the identity mapping is also included as one of the branches in the multi-branch framework. Through learnable coefficients for each branch, the structure effectively decouples linear and nonlinear features and adaptively balances the nonlinear feature expressiveness and linear feature retention capabilities across different layers of the network.

When selecting Identity, ReLU, PReLU, and HardSwish-type activation functions for the power function as the components of RepAct I multi-branch structure, the activation function representation of RepAct I during the training phase is expressed as equation 2.
\begin{equation}
    X' = a_0Identity(X)+a_1ReLU(X)+a_2PReLU(X)+a_3HardSwish(x)
\end{equation}

During the network training phase, the weight factors of each branch are learnable adaptive parameters. During the network inference phase, they are re-parameterized and merged as static constants. Thus, the RepAct I activation function at the inference stage can be simplified to equation 3, restoring it to a single-branch power function form with segmented sections. In the combination, the single-branch segment interval of RepAct I after re-parametrization becomes the union of branch segment intervals, and $\alpha_{n}$ is re-parametrized as $\delta_{n}$ to serve as a static network parameter.
\begin{equation}
\begin{array}{l}
X^{\prime}=\operatorname{Rep} A c t(X)=\left\{\begin{array}{l}
x^{*}\left(\alpha_{0}+\alpha_{1}+\alpha_{2}+\alpha_{3}\right), 3 \leq x \\
x^{*} x^{*}\left(\alpha_{3} / 6\right)+x^{*}\left(\alpha_{0}+\alpha_{1}+\alpha_{2}+\alpha_{3} / 2\right), 0 \leq x<3 \\
x^{*} x^{*}\left(\alpha_{3} / 6\right)+x^{*}\left(\alpha_{0}+0+\left(\text { PReLU.T* } \alpha_{2}\right)+\alpha_{3} / 2\right),-3 \leq x<0 \\
x^{*}\left(\alpha_{0}+0+\left(\text { PReLU.T* } \alpha_{2}\right)+0\right), x<-3
\end{array}\right. \\
=\left\{\begin{array}{l}
x^{*} \delta_{1}, 3 \leq x \\
x^{*} x^{*} \delta_{2}+x^{*} \delta_{3}, 0 \leq x<3 \\
x^{*} x^{*} \delta_{2}+x^{*} \delta_{4},-3 \leq x<0 \\
x^{*} \delta_{5}, x<-3
\end{array}, \text { In the formula }\left(\begin{array}{l}
\delta_{1}=\alpha_{0}+\alpha_{1}+\alpha_{2}+\alpha_{3} \\
\delta_{2}=\alpha_{3} / 6 \\
\delta_{3}=\alpha_{0}+\alpha_{1}+\alpha_{2}+\alpha_{5} / 2 \\
\delta_{4}=\alpha_{0}+0+\left(\text { PReLU.T } * \alpha_{2}\right)+\alpha_{3} / 2 \\
\delta_{5}=\alpha_{0}+0+\left(\text { PReLU.T } * \alpha_{2}\right)+0
\end{array}\right)\right.
\end{array}
\end{equation}

After re-parameterization of RepAct I, the number of model parameters has increased by only five parameters in each layer implementing RepAct. Within the various branches of RepAct , the maximum computational complexity function (O(RepAct\_n) corresponds to HardSwish, as shown in equation 4. Compared to the maximum complexity of (O(RepAct\_n), the computation of RepAct is increased by only one if operation and two multiplication operations.
\begin{equation}
\text { HardSwish }(X)=\left\{\begin{array}{l}
x, 3 \leq x \\
x^{*} x^{*} \frac{1}{6}+x^{*} \frac{1}{2},-3 \leq x<3 \\
0, x<-3
\end{array}, \text { RepAct } \mathrm{I}(X)=\left\{\begin{array}{l}
x * \delta_{1}, 3 \leq x \\
x^{*} x^{*} \delta_{2}+x^{*} \delta_{3}, 0 \leq x<3 \\
x^{*} x^{*} \delta_{2}+x^{*} \delta_{4},-3 \leq x<0 \\
x^{*} \delta_{5}, x<-3
\end{array}\right.\right.
\end{equation}

During the deployment phase, in GPUs or AI-accelerated edge devices, activation functions are often fused with convolution computations to reduce frequent memory access. Convolution processing is typically parallel or partially parallel, so for the fused segmented power series class activation functions, the bottleneck of inference speed is the segment with the maximum computation among all segments. After re-parameterization of RepAct I, the computational complexity of the segment with the maximum computation remains as $O(ax^{2}+bx)$, consistent with max(O($RepAct_n$), which allows the approximation $O(RepAct) \approx max(O(RepAct_n))$.

During the initialization of RepAct I's branch weight factors, a simple approach is to evenly distribute the branch weights so that their sum equals one. This ensures that during the initial stages of network training, the sum of features and gradients are not too small to vanish, enabling the effective transmission of different features across each branch. Following initialization, the activation function of RepAct I and its first-order derivative are as illustrated in figures 2(a), (b), and (c), respectively showing each branch's activation mapping and first-order derivatives within the [0,1] range.
\begin{figure}[h]
\centering
    
    \begin{subfigure}{0.3\textwidth}
        \includegraphics[width=\textwidth]{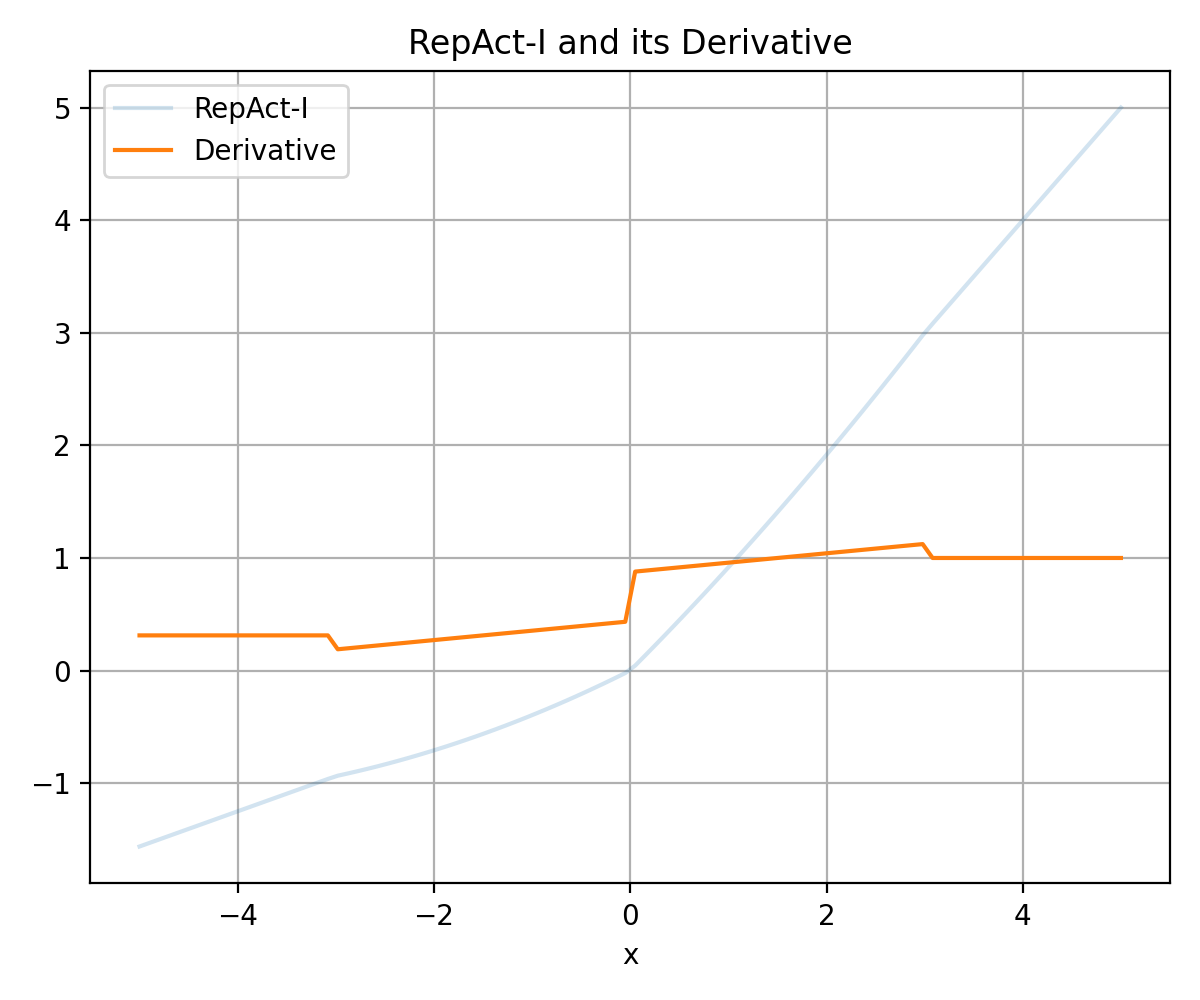}
        \label{fig:sub1}
    \end{subfigure}
    \hfill
    \begin{subfigure}{0.3\textwidth}
        \includegraphics[width=\textwidth]{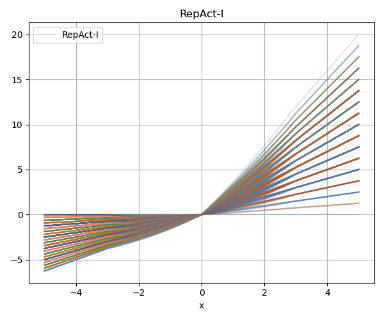}
        \label{fig:sub2}
    \end{subfigure}
    \hfill
    \begin{subfigure}{0.3\textwidth}
        \includegraphics[width=\textwidth]{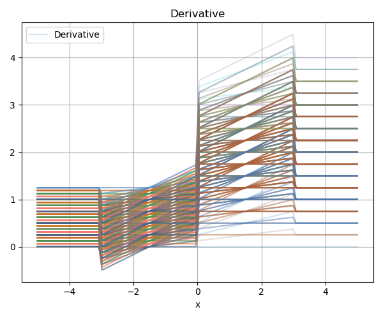}
        \label{fig:sub3}
    \end{subfigure}
\caption{RepAct I Activation function and its first derivative(a) Initialize the RepAct on average(b) $\alpha_n$[0-1] The RepAct I after the fluctuation(c) $\alpha_n$[0-1] The RepAct I reciprocal image}
\label{fig2:main}
\end{figure}
\subsection{Adaptive scaling of forward feature and reverse gradient}
\begin{figure}[h]
    \centering
    \includegraphics[width=0.95\textwidth]{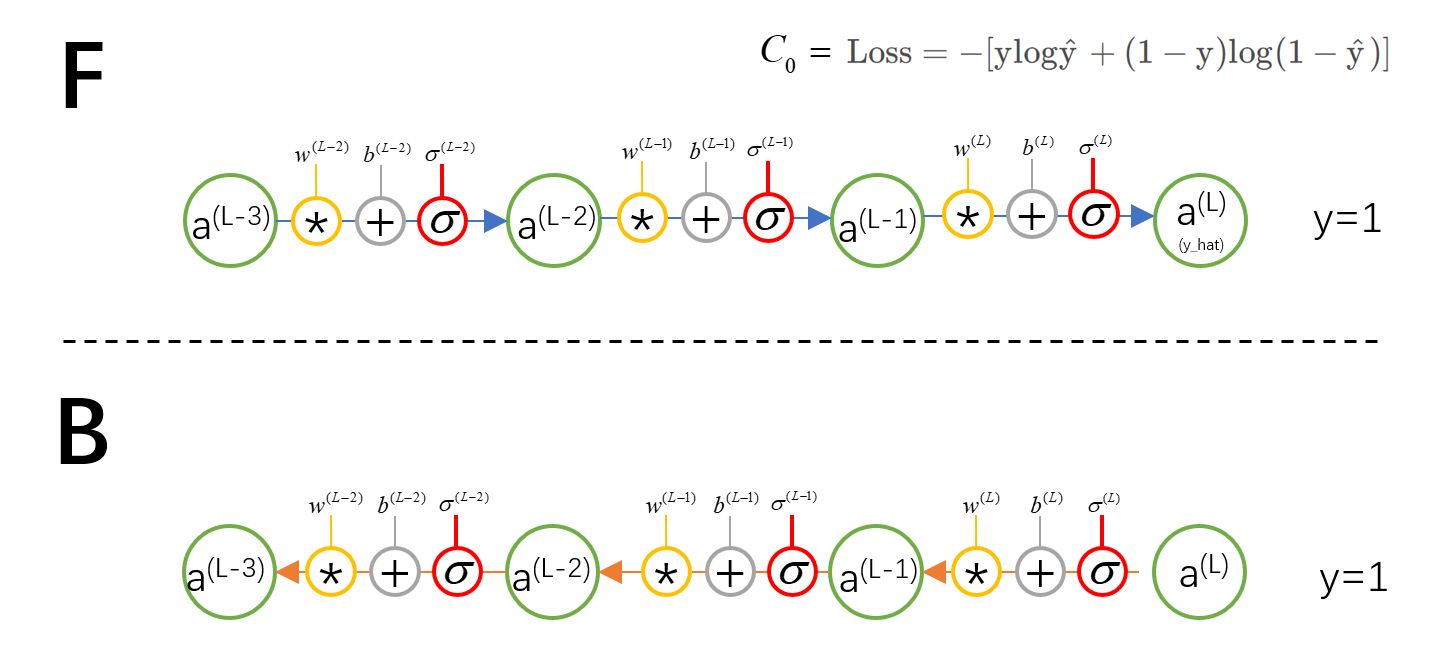}
    \caption{Forward and backward diagram of neural network}
    \label{fig:enter-label}
\end{figure}

Forward reasoning:
\begin{equation}
    a^{\left(L\right)}=\sigma^{\left(L\right)}z^{\left(L\right)}
\end{equation}
\begin{equation}
    z^{\left(L\right)}=w^{\left(L\right)}a^{\left(L-1\right)}+b^{\left(L\right)}
\end{equation}
\begin{equation}
   C_0=\text{LOSS}=-\left[ y \log(\hat{y}) + (1-y) \log(1-\hat{y}) \right]
\end{equation}
($a^{\left(L\right)}$ is the active value of the current layer,$\sigma^{\left(L\right)}$ is the current layer activation function,$w^{\left(L\right)}$ and $b^{\left(L\right)}$ they are weight and bias,$C_0$ is a loss function)

$\frac{\partial C_0}{\partial w^{(L)}} $ Inverse gradient calculation:
\begin{equation}
    \frac{\partial C_0}{\partial w^{(L)}} =\frac{\partial z^{{L}}}{\partial w^{{L}}}\frac{\partial a^{{L}}}{\partial z^{{L}}} \frac{\partial C_0^{{L}}}{\partial a^{{L}}}
\end{equation}
\begin{equation}
    \frac{\partial C_0}{\partial a^{(L)}} = loss\_derivative 
\end{equation}
\begin{equation}
    \frac{\partial a}{\partial z^{(L)}} = \sigma ^{(L)}{}' (z^{(L)})
\end{equation}
\begin{equation}
    \frac{\partial z{(L)}}{\partial w^{(L)}} = a^{(L-1)}
\end{equation}

During the calculation of gradients for $\frac{\partial C_0}{\partial w^{(L)}} $ at layer L, the scaling effect of $\sigma^{(L)}$ directly influences the derivative of $\frac{\partial a^{(L)}}{\partial z^{(L)}}$.In the forward process, there is an indirect impact on $\frac{\partial C_0}{\partial a^{(L)}} $ through the effect on $a^{(L)}$.As $a^{(L-1)}$ is subject to scaling through $\sigma^{(L-1)}$ in the forward direction,$\frac{\partial z^{(L)}}{\partial w^{(L)}} $is also indirectly affected by $\sigma^{(L-1)}$.Therefore, the scaling action of the activation function will affect the feature values being propagated and the gradients during the backward propagation for the current layer as well as the subsequent layers, forming a hierarchical gradient scaling mechanism.
\begin{figure}[h]
    \centering
    \includegraphics[width=0.95\textwidth]{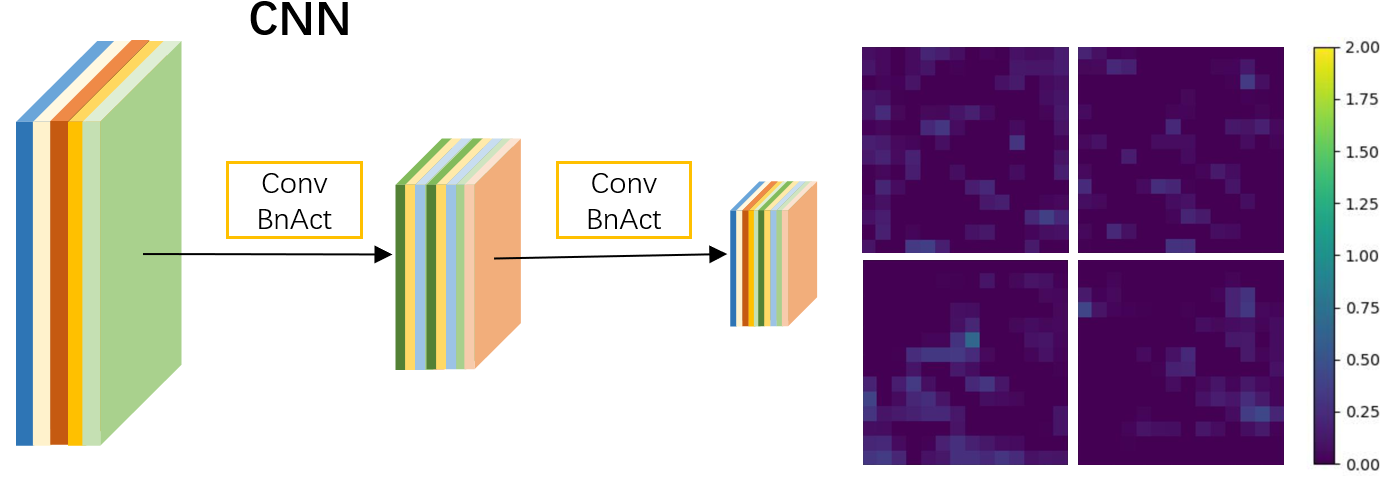}
    \caption{CNN network structure and feature map sparse activation heat map}
    \label{fig:enter-label}
\end{figure}

In CNNs, the scaling effect of activation functions on features and gradients is similar. As shown in Figure 4, each convolutional layer is typically followed by a Batch Normalization (BN) layer \cite{12,37,38}, which adaptively normalizes forward features and backward gradients. The BN layer ensures that the scaling effect of the activation function in the current layer does not cause an explosion or vanishing of features or gradients in more distant feature or gradient updates \cite{38}. Hence, when updating the weights of each layer in a CNN, the gradient scaling caused by the activation function is only significantly effective for that particular layer. For the shallow layers of the network, after BN processing, they carry back more category information to the shallow layers; therefore, RepAct can enhance the learning ability of various lightweight CNNs.

On the other hand, CNN feature maps usually exhibit sparse activations \cite{16}, suggesting that after CNN training is complete, the network can be compressed and pruned in different dimensions \cite{39,40,41,42}. This indicates that sparse activations in CNNs lead to redundant network weights, preventing the full utilization of the total weight information capacity for task learning. RepAct continuously adjusts the weight factors of each channel branch, adaptively capturing and scaling features during forward inference and scaling the gradient magnitude in each layer during backward propagation. Thus, RepAct forms a gradient scaler that not only adapts to network layer structures but also to the inter-layer feature distribution, which is more conducive to the network's efficient use of its information capacity for learning.

In Figure 5, within RepAct, the Identity and ReLU functions represent linear and nonlinear mappings, respectively, each with their own branch weight factors. During the backward propagation of the network, after re-parameterization of RepAct to a single branch, the partial derivative of the weight coefficient $k_1$ for $X>0$ will be further decoupled. Utilizing the prior segmentation experience of Identity and ReLU, an update is formulated for the decoupled linear and nonlinear mappings, which together affect $k_1$.
\begin{figure}
    \centering
    \includegraphics[width=0.95\textwidth]{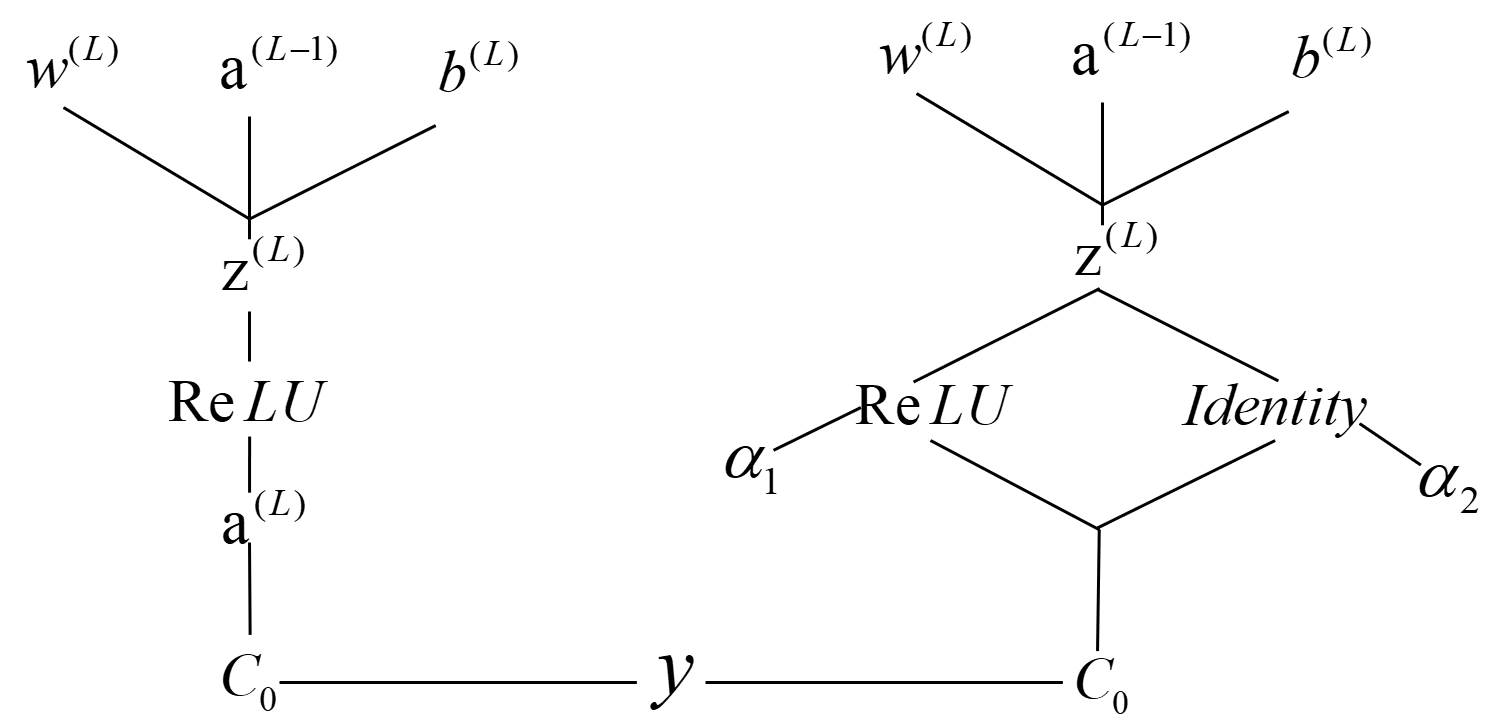}
    \caption{Schematic diagram of ReLU and RepAct (Identity and ReLU) backpropagation chain}
    \label{fig:enter-label}
\end{figure}
\begin{equation}
    \operatorname Repact=\left\{\begin{array}{l}
k_{1} x, x>0 \\
k_{2} x, x<0
\end{array}, k_{1}=\alpha_{1}+\alpha_{2}, k_{2}=\alpha_{2}\right.
\end{equation}
\begin{equation}
    k_1 = k_1-\ell(\partial a_1 +\partial a_2),k_2 = k_2-\ell(\partial a_2)
\end{equation}
\subsection{RepAct II competition and III cooperation}
\subsubsection{Repact-II-Softmax - Competition}
\begin{figure}[h]
    \centering
    \includegraphics[width=0.95\textwidth]{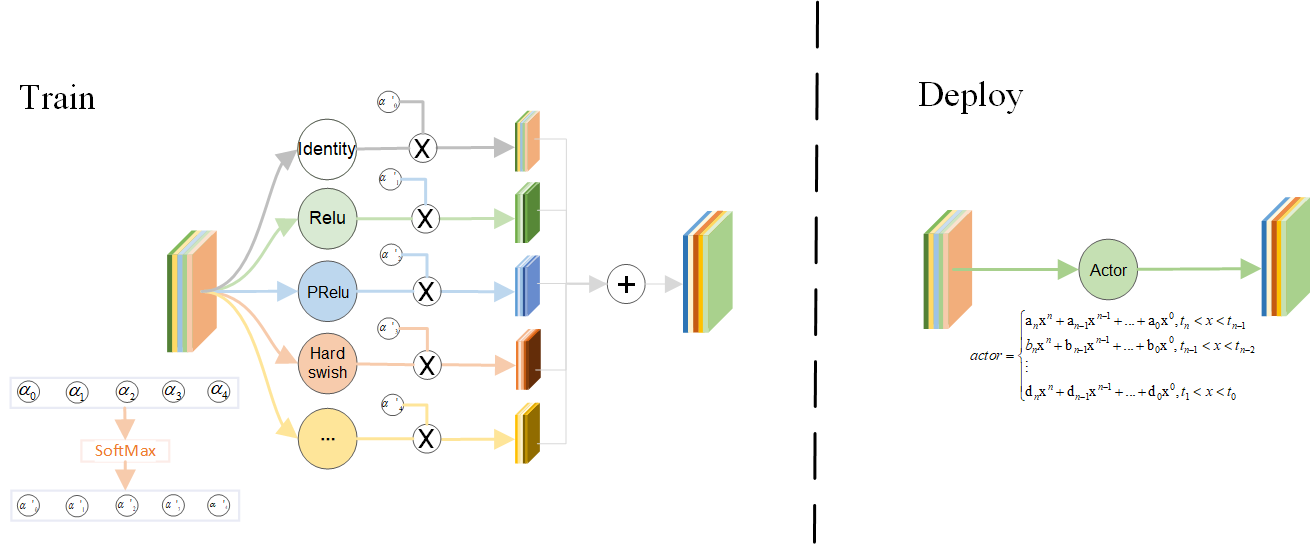}
    \caption{RepACt-II-Softmax-competition training multi-branch and inference single-branch structure}
    \label{fig:enter-label}
\end{figure}

When RepAct I $\sum a_i>1$, feature information passing through the RepAct layer is magnified during the forward propagation, amplifying the feature itself. Similarly, during backpropagation, the feature gradient is also magnified due to the chain rule, which can facilitate the learning of network parameters. However, this may also lead to the network easily capturing noise and potentially causing overfitting. Therefore, it becomes necessary to impose constraints on the adaptive weight factors of the branches.

We further propose the RepAct II-Softmax model, which features a degradable soft gating adaptive factor. The adaptive factors $\alpha$ of each branch are mapped through a Softmax function, resulting in $\sum {\alpha}' =1$ after the mapping for each branch. This approach reduces the magnifying effect of the activation layer, creating a limited-resource competition among the branch activation functions, which in turn helps reduce the network's tendency to overfit.
\begin{equation}
    {\alpha}' = \frac{e^{\alpha _i}}{\sum_{i=1}^{K}e^{a_i}},\sum {\alpha}'=1  
\end{equation}
($\alpha_i$is the branch i weight before the mapping,${\alpha_i}'$is the weight of branch i after mapping)

During the inference stage,${\alpha}'$ is retained after the Softmax mapping, and the system reverts back to the RepAct I form. This process involves re-parameterization of the various branches to restore the single-branch activation function, consistent with equation (3).
\subsubsection{RepAct III-BN - Cooperation}
\begin{figure}[h]
    \centering
    \includegraphics[width=0.95\textwidth]{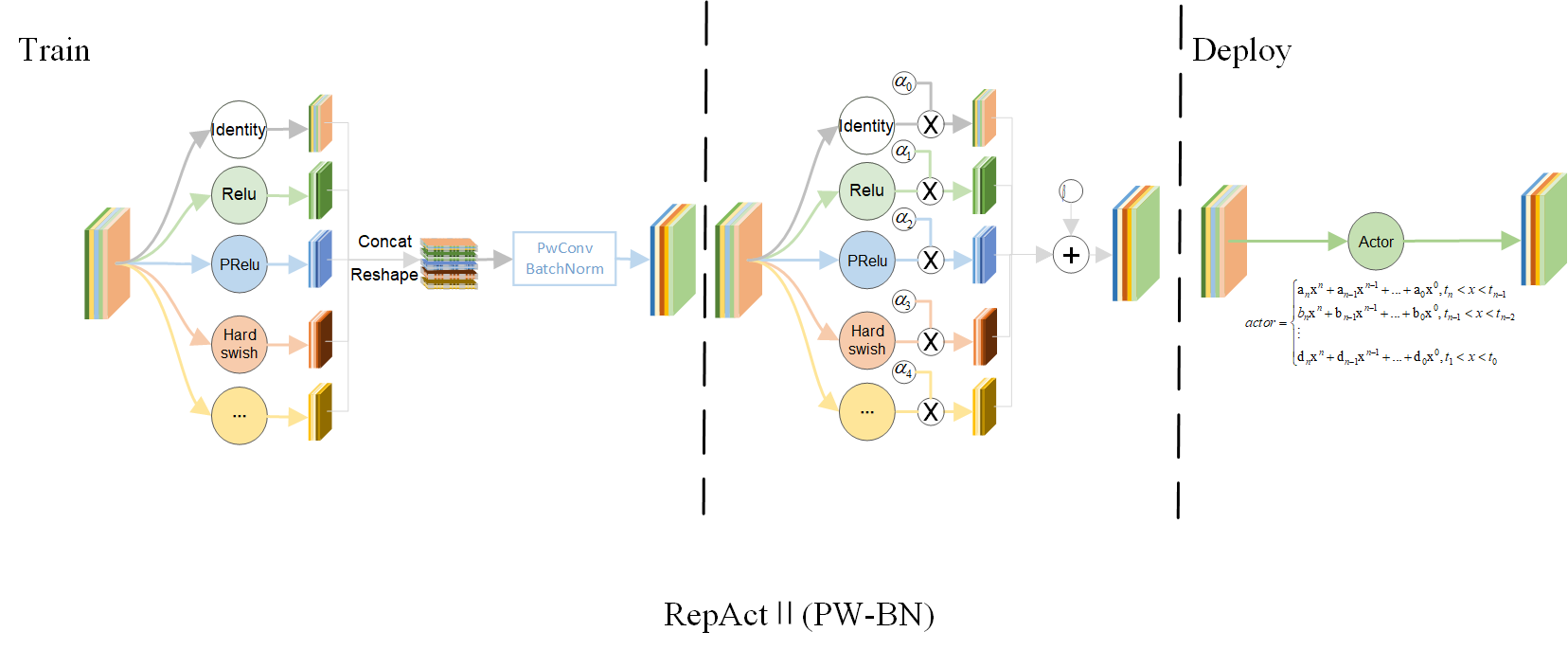}
    \caption{Training multi-branch and inference single-branch structures of RepAct III-BN-cooperation}
    \label{fig:enter-label}
\end{figure}

Intuitively, we expect that aside from the backward gradient descent adaptive adjustments, the single-branch factor $\alpha_i$ should also cooperate synergistically with the factors of other branches to modulate the branch weights, allowing the fused feature maps after branching to aggregate more efficient feature information. However, we do not wish to introduce additional auxiliary memory access operations during the inference stage, as this would slow down network inference speeds.

Consequently, we further propose the RepAct III-BN cooperative type, which possesses degradable global information. This design allows the branch weight factors to make use of the global information from the feature maps after branch fusion to readjust during the BN fusion stage.

RepAct III-BN first overlays the feature maps activated by each branch, resulting in a fused feature map. In the second stage, this fused feature map is treated as an integrated whole and fed into a single-channel BN layer where it undergoes an affine transformation using the global information of the mean and variance of the fused feature map and the learnable parameters. The final feature map after the BN layer is then the final output feature map of RepAct III-BN.

During the branch fusion phase, the linear properties of the BN layer are first leveraged to merge it with the factors from each branch of RepAct. The fusion of RepAct III-BN is formalized as equation 15. This means that in the inference phase, RepAct III-BN degrades to the form shown in Figure 7, RepAct III-BN (b), where each branch factor ${\alpha_i}'$ after BN linear degradation is as described in equation 16. After the degradation of BN, the branch factor ${\alpha_i}'$ of RepAct III-BN reverts to the form of RepAct I. The factor ${\beta}'$complements or is amalgamated into the segments as a zero-power coefficient after the fusion of RepAct I branches, completing the training structure from multi-branch structure to single-branch inference of RepAct III-BN. The re-parameterization post-training is as depicted in Figure 7, RepAct III-BN (c). As such, RepAct III-BN, through two stages of re-parameterization degradation, enables each branch's adaptive factor to possess the global feature information of the post-fusion feature maps.
\begin{equation}
    \begin{aligned}
        y_{final}=BN(a_0x_0+a_1x_1+...+a_nx_n)=\gamma (\frac{x_N-u}{\sqrt{ \sigma^{2}+1e^{-6}} })+\beta \\
    = \frac{\gamma x_N}{\sqrt[]{\sigma ^{2}+1e^{-6}} }+(\beta -\frac{\gamma u}{\sqrt[]{\sigma ^{2}+1e^{-6}} } )=\varepsilon x_N+{\beta}'
    \end{aligned}    
\end{equation}
In the equation 15,$a_0x_0+a_1x_1+...+a_nx_n=x_n,;\varepsilon =\frac{\gamma }{\sqrt[]{\sigma ^{2}+1e^{-6}}} ;{\beta}' = \beta -\frac{\gamma u}{\sqrt[]{\sigma ^{2}+1e^{-6}} }$
\begin{equation}
    {\alpha}'_i = \varepsilon {\alpha}_i =\frac{\gamma }{\sqrt[]{\sigma ^{2}+1e^{-6}} }{\alpha}_i
\end{equation}
\begin{figure}[h]
    \centering
    \includegraphics[width=0.95\textwidth]{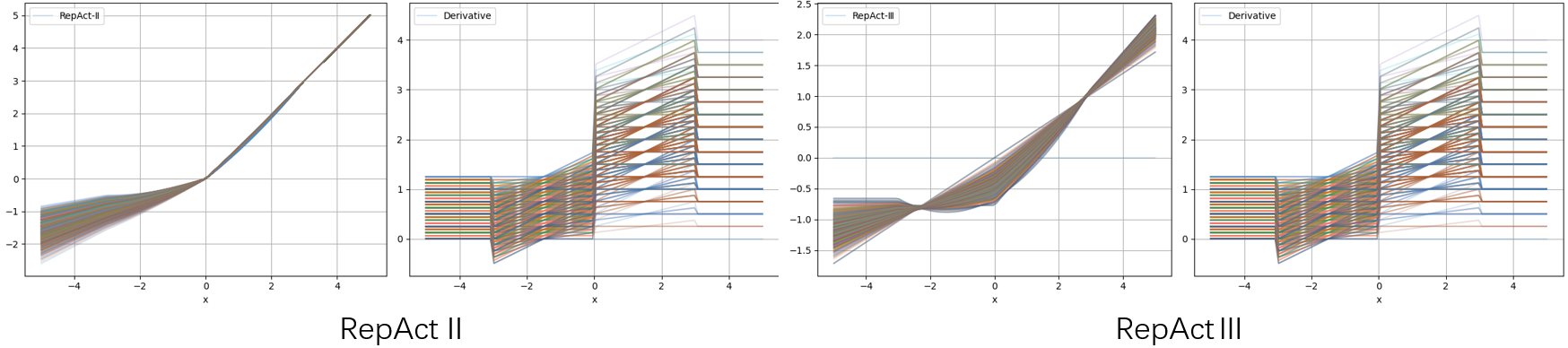}
    \caption{Visualization of RepAct-II-SoftMax-competition and RepAct-III-BN-cooperation activation functions and their derivatives}
    \label{fig:enter-label}
\end{figure}
\section{Experiments}

To comprehensively evaluate the performance of RepAct in classical lightweight network architectures and various tasks, we conducted a thorough validation across multiple datasets, including Imagenet100, Cifar100, VOC12-Detect, and VOC12-Segment. With the use of GradCAM for visualization, we intuitively demonstrated the enhancement of the feature attention areas in lightweight networks by RepAct. This provided an in-depth analysis of the working mechanism behind the enhancement of network learning capabilities from a gradient perspective.

\subsection{Image Classification}

Utilizing the ImageNet100 dataset (composed of 100 categories extracted from the ImageNet2012-1k) \cite{29} and the Cifar100 dataset, we conducted ablation experiments on classic lightweight networks \cite{12,37} and small ViT models \cite{43,44} by simply replacing the original network activation functions with the RepAct series. For the training, we used image sizes of 224 pixels (ImageNet100) and 32 pixels (Cifar100) as inputs, to perform network learning using the conventional image classification training approach (detailed settings of the training hyperparameters can be reviewed in the appendices, with all tasks having a fixed random seed of 0).

\begin{center}
\noindent Table 1: Original activation network and RepAct series network Top-1acc on ImageNet100. Correlation table
\resizebox{1.0\linewidth}{!}{
\begin{tabular}{*{14}{c}}
  \toprule
  ImageNet100/Top-1 acc  &Baseline&PReLu&ACONC&RepAct-I&RepAct-II&RepAct-III\\
  \midrule
  ShuffleNetV2x0.5    &73.0(R)&+3.64&+3.96&\textcolor{red}{+5.54}&+4.04&+4.84 \\
  ShuffleNetV2x1.0   &80.8(R)&+0.82&+0.78&\textcolor{red}{+1.92}&+1.52&+0.7 \\
  MobileNetV3-Small    &72.88(H)&+6.88&+6.34&+6.9&+6.18&\textcolor{red}{+7.92} \\
   MobileNetV3-Large    &79.78(H)&+3.8&+3.22&\textcolor{red}{+4.38}&+4.36&\textcolor{red}{+4.38} \\
SwinTinyPatch4Window7  &70.14(G)&/&/&\textcolor{red}{+9.54}&+5.98&+8.1 \\
  VitVasePatch16   &58.92(G)&/&/&+3.74&\textcolor{red}{+5.66}&-12.8 \\

  \bottomrule
\end{tabular}
}
\end{center}

\begin{center}
\noindent Table 2: Original activation network vs. RepAct series network Top-1acc on Cifar100. Correlation table
\resizebox{1.0\linewidth}{!}{
\begin{tabular}{*{14}{c}}
  \toprule
  Cifar100/ Top-1 acc  &Baseline&PReLu&ACONC&RepAct-I&RepAct-II&RepAct-III\\
  \midrule
  ShuffleNetV2x0.5    &66.71(R)&+0.72&-0.58&+0.04&\textcolor{red}{+1.05}&-1.32\\
  ShuffleNetV2x1.0   &70.82(R)&+0.36&+0.59&+0.43&\textcolor{red}{+0.86}&-0.38 \\
  MobileNetV3-Small    &67.35(H)&+1.77&+3.96&\textcolor{red}{+5.71}&+2.32&+3.52 \\
   MobileNetV3-Large    &70.85(H)&+2.23&+2.25&+3.74&+0.37&\textcolor{red}{+4.04} \\
  \bottomrule
\end{tabular}
}
\end{center}

\begin{center}
\noindent Table 3: Activation functions Top-1acc of MobileNetV3-Small in Cifar100. Correlation table
\resizebox{1.0\linewidth}{!}{
\begin{tabular}{*{14}{c}}
  \toprule
    Act&Top-1 acc.&Act&Top-1 acc.&Act&Top-1 acc.&Act&Top-1 acc.\\
  \midrule
  HardSwish&67.35&ReLU&+1.46&Swish&+2.72&ELU&-1.83\\
  RepAct-I&\textcolor{red}{+5.71}&LReLU&+2.98&Mish&+2.05&SELU&-2.37\\
  RepActII&+2.32&PReLU&+1.77&FReLU&+3.56&GELU&+3.62\\
  RepAct-III&+3.52&AReLU&+2.00&DYReLU&+0.94&CELU&-1.75\\
  Identity&-35.90&Softplus&-14.42&ACONC&+3.96&SiLU&+1.90 \\
  \bottomrule
\end{tabular}
}
\end{center}

\begin{center}
\noindent Table 4: Original activation network vs. RepAct series network Top-1acc on Cifar10. Correlation table
\resizebox{1.0\linewidth}{!}{
\begin{tabular}{*{14}{c}}
  \toprule
  Cifar100/ Top-1 acc  &Baseline&PReLu&ACONC&RepAct-I&RepAct-II&RepAct-III\\
  \midrule
  ShuffleNetV2x0.5    &90.54(R)&+0.27&-0.22&\textcolor{red}{+0.62}&+0.39&+0.33\\
  ShuffleNetV2x1.0   &92.76(R)&+0.28&-0.39&\textcolor{red}{+0.29}&+0.15&+0.17 \\
  MobileNetV3-Small    &91.22(H)&+0.92&+0.26&\textcolor{red}{+1.11}&\textcolor{red}{+1.11}&+0.92 \\
   MobileNetV3-Large    &92.68(H)&+1.56&+1.09&+1.31&\textcolor{red}{+1.7}&+1.44 \\
  \bottomrule
\end{tabular}
}
\end{center}

\begin{center}
\noindent Table 5: MobileNetV3-Small activation functions Top-1acc in Cifar10. Correlation table
\resizebox{1.0\linewidth}{!}{
\begin{tabular}{*{14}{c}}
  \toprule
    Act&Top-1 acc.&Act&Top-1 acc.&Act&Top-1 acc.&Act&Top-1 acc.\\
  \midrule
  HardSwish&91.22&ReLU&+0.93&Swish&+0.74&ELU&-0.76\\
  RepAct-I&\textcolor{red}{+1.11}&LReLU&+1.05&Mish&+0.35&SELU&-1.39\\
  RepActII&\textcolor{red}{+1.09}&PReLU&+0.92&FReLU&\textcolor{red}{+1.29}&GELU&+0.9\\
  RepAct-III&+0.92&AReLU&+0.99&DYReLU&+0.78&CELU&-0.76\\
  Identity&-31.06&Softplus&-2.46&ACONC&+0.26&SiLU&+0.13 \\
  \bottomrule
\end{tabular}
}
\end{center}

In Table 1, various classic lightweight CNN networks have shown a significant improvement in classification accuracy on the ImageNet100 dataset, thanks to the RepAct’s in-training adaptive combination of branch features without introducing additional memory operations for feature map access during the inference phase. This effect is especially pronounced for lighter networks like ShuffleNetV2x0.5 and MobileNetV3-Small, which saw improvements of 5.54\% and 7.92\% in Top-1 accuracy, respectively. It is observable that RepAct I, as compared to the cooperative RepAct II and competitive RepAct III models, performs more effectively in data-rich classification tasks with unconstrained adaptive branch coefficients. In Table 2, on the Cifar100 dataset, each adaptive activation function shows less improvement in ShuffleNetV2 than in MobilenetV3, suggesting that in classifications of less difficult datasets where the network can fully utilize its own information capacity, the RepAct II-Softmax type with degradable soft gating adaptive factors can reduce network overfitting. In Tables 3, RepAct-I outperforms other mainstream activation functions, whereas RepAct-III achieves accuracy levels comparable to those of activation functions like FReLU—which requires additional memory computation—or those with higher exponential computational complexity (like GELU, ACONC), solely by using a computation of the power function form twice. In Tables 4 and 5, RepAct also shows good performance on the Cifar10 classification task, closely matching the additional computation required by FReLU.To sum up, RepAct significantly improves the learning ability of lightweight networks in the training of various tasks without increasing the reasoning burden of lightweight networks.

The more the parameters in lightweight models are reduced, the total capacity from a model capacity perspective diminishes, causing a decrease in the breadth and complexity of data distributions the model can adapt to. From the perspective of model learning, it is expected that a model should utilize its total information capacity as fully as possible for learning within tasks, but a decrease in parameter quantity can result in a decline in feature extraction capability during the training phase. If the gradients backpropagated are ineffective in updating network weights with category information-rich feature gradients, the model can't efficiently use its information capacity, leading to an overall performance decline—that is, the model cannot perform efficient learning. RepAct enhances model training complexity and feature extraction capability through adaptive multi-branch training, enabling adaptive feature information scaling. In the backward propagation stage of network training, the selective feature amplification due to RepAct's layer granularity in forward inference is reversed during gradient backpropagation, magnifying the gradient chain. This allows network weights to be updated more easily focused on these features, making full use of the total information capacity for efficient task learning.

We also verified on the minimum version ViT and Swin Transformer that due to RepAct's ability to efficiently learn from the adaptive branch selection of different layers and forward feature information scaling, lightweight Transformers also significantly improved their classification accuracy under the same training conditions. This showcases the broad adaptability of RepAct to enhance the learning capability of lightweight networks.
\begin{figure}[h]
    \centering
    \includegraphics[width=0.95\textwidth]{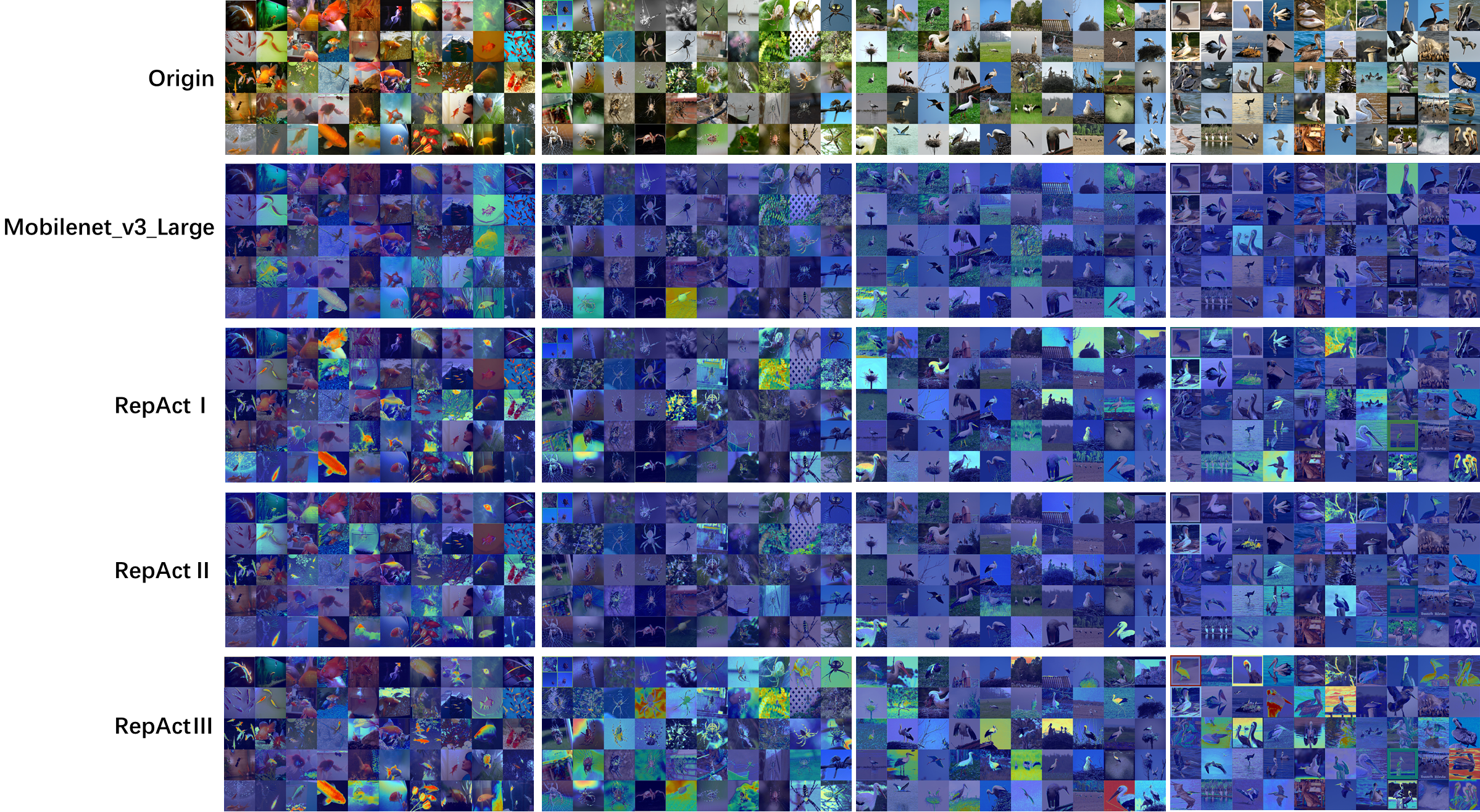}
    \caption{Comparison of visual effects of MobileNetV3-Large-First-Feature GradCAM}
    \label{fig:enter-label}
\end{figure}

\begin{figure}[h]
    \centering
    \includegraphics[width=0.95\textwidth]{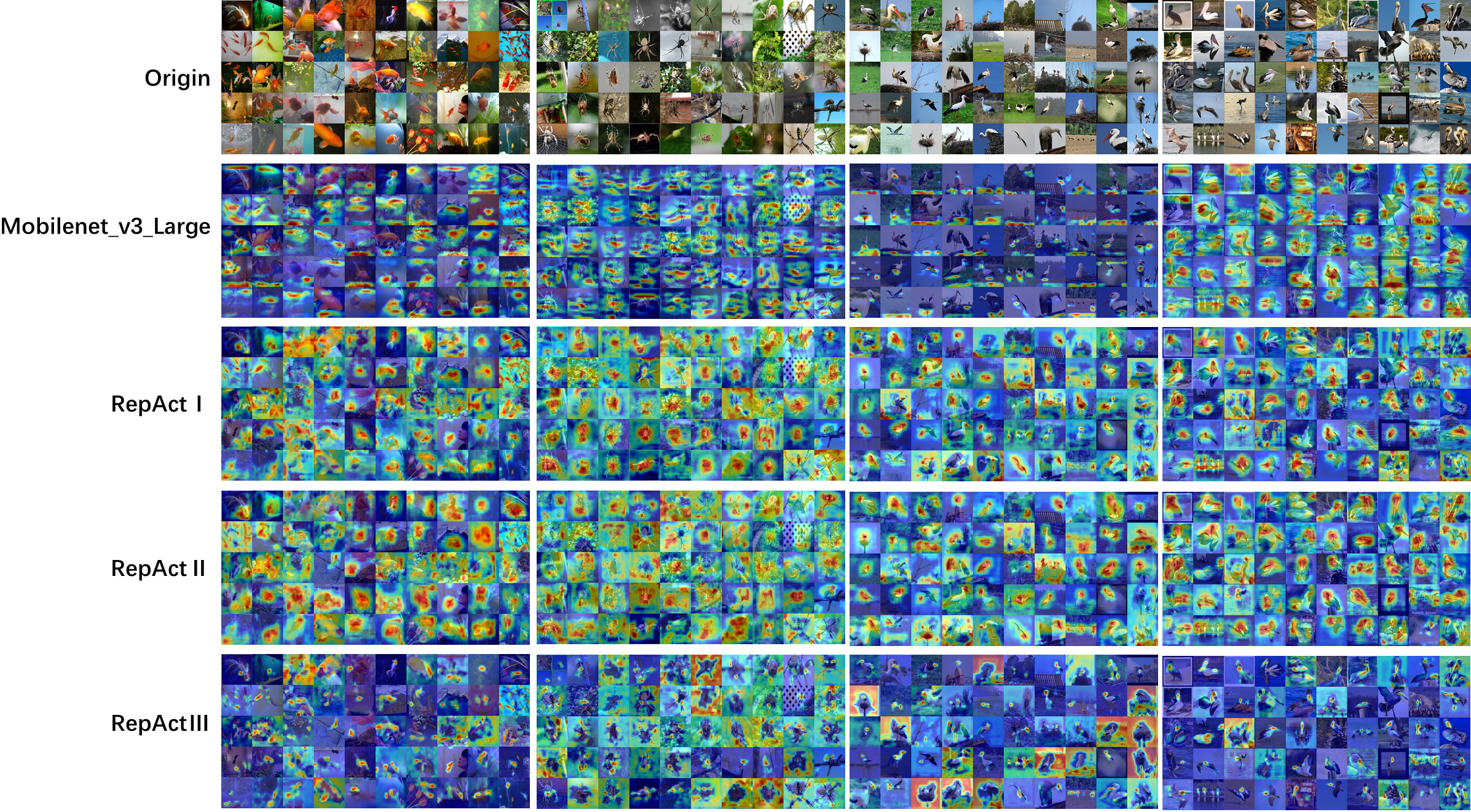}
    \caption{Comparison of visual effects of MobileNetV3-Large-Last-Feature GradCAM}
    \label{fig:enter-label}
\end{figure}

In the GradCAM \cite{49} comparative visualization shown in Figure 9, the texture information extracted by the shallow layers of the RepAct series networks aligns more closely with the local texture information required for the specific classification category, and the area of attention is pronounced. This concurs with the magnification of effective information features during the forward process of RepAct, as well as the efficient scaling and back-propagation of gradients bearing category information during the backward propagation. In the backward chain rule derivation, deeper network gradients transmit more category information back to the shallower network layers. The shallow layers learn to capture texture features with more category information, making full use of the total model capacity in task and data learning.

In Figure 10, the original network focuses only on local features of class objects during deep convolutional layers, and the network's attention heat maps often display some offset from the actual position of the class objects; RepAct I and II show superior capabilities in semantic-level attention to the entire category of the class, significantly focusing on the entire class object and also paying some ancillary attention to the pixel environment surrounding the class object. RepAct III provides more accurate attention to local features associated with the category, such as high attention to bird beaks and fish bodies. RepAct III, with its global degradable information, may lead to overfitting background information.
\subsubsection{RepAct and gradient visual analysis}
\begin{figure}[h]
    \centering
    \includegraphics[width=0.95\textwidth]{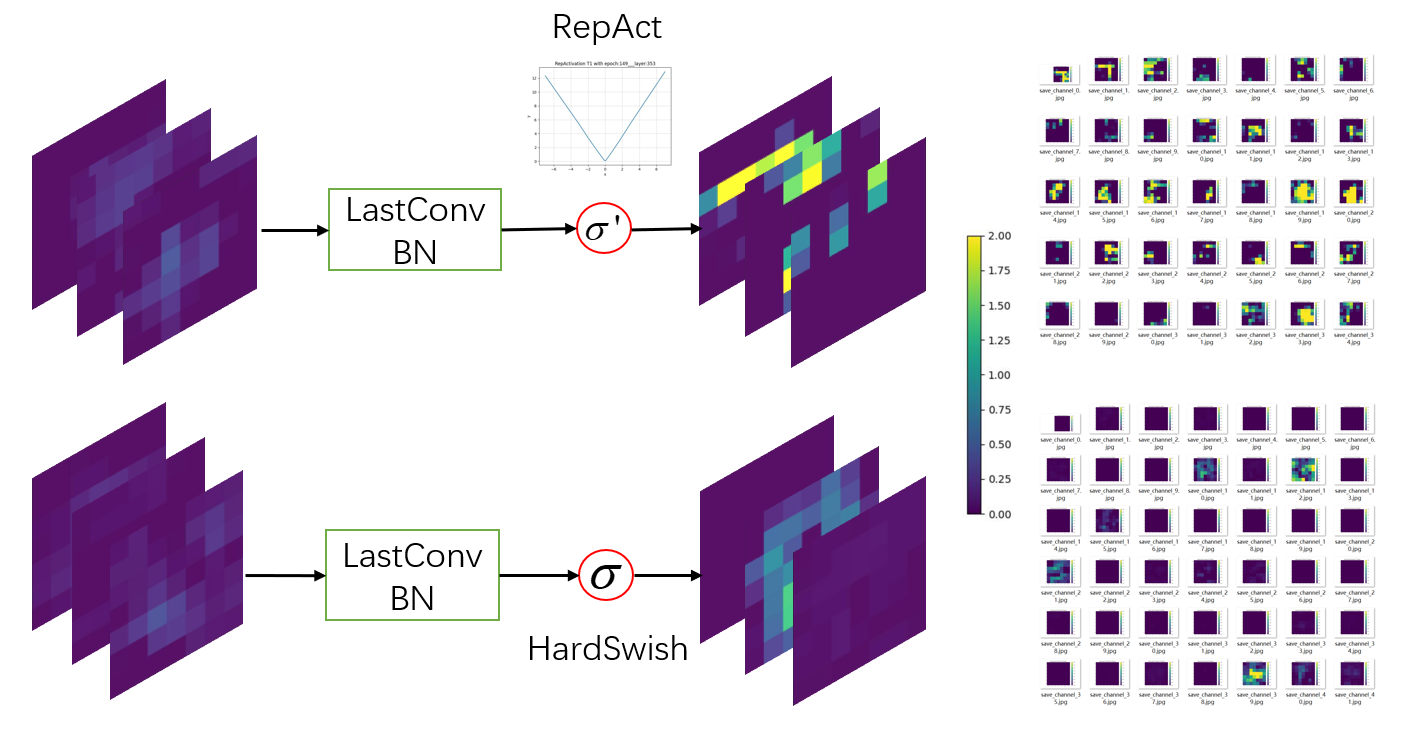}
    \caption{Feature map thermal map visualization of RepAct I and HardSwish in MobileNetV3-Large after ImageNet100 training before and after passing the final convolutional layer}
    \label{fig:enter-label}
\end{figure}

In Figure 11, at the last layer of convolution, feature map activations through HardSwish are relatively sparse, with only a small number of feature map channels being activated. Through the adaptive V-shaped feature maps of RepAct I, more channel feature maps are activated. During the weight gradient calculation of the last convolutional layer (LastConv), the gradients are magnified by RepAct-I. Furthermore, the gradient calculation for layers above LastConv will depend on the gradient information from that layer, making it easier for loss gradients containing category information to backpropagate to the shallower layers. In the gradient computation of the subsequent classifier weights, the output feature maps of the LastConv layer are cross-correlated with the gradient matrix of the loss function derivatives. As feature maps are more activated, it promotes easier updating of classifier weights, endowing the network with stronger feature learning and feature transmission capabilities.
\begin{figure}[ht!]
    \centering
    \begin{subfigure}[b]{0.8\textwidth}
        \centering
        \includegraphics[width=\textwidth]{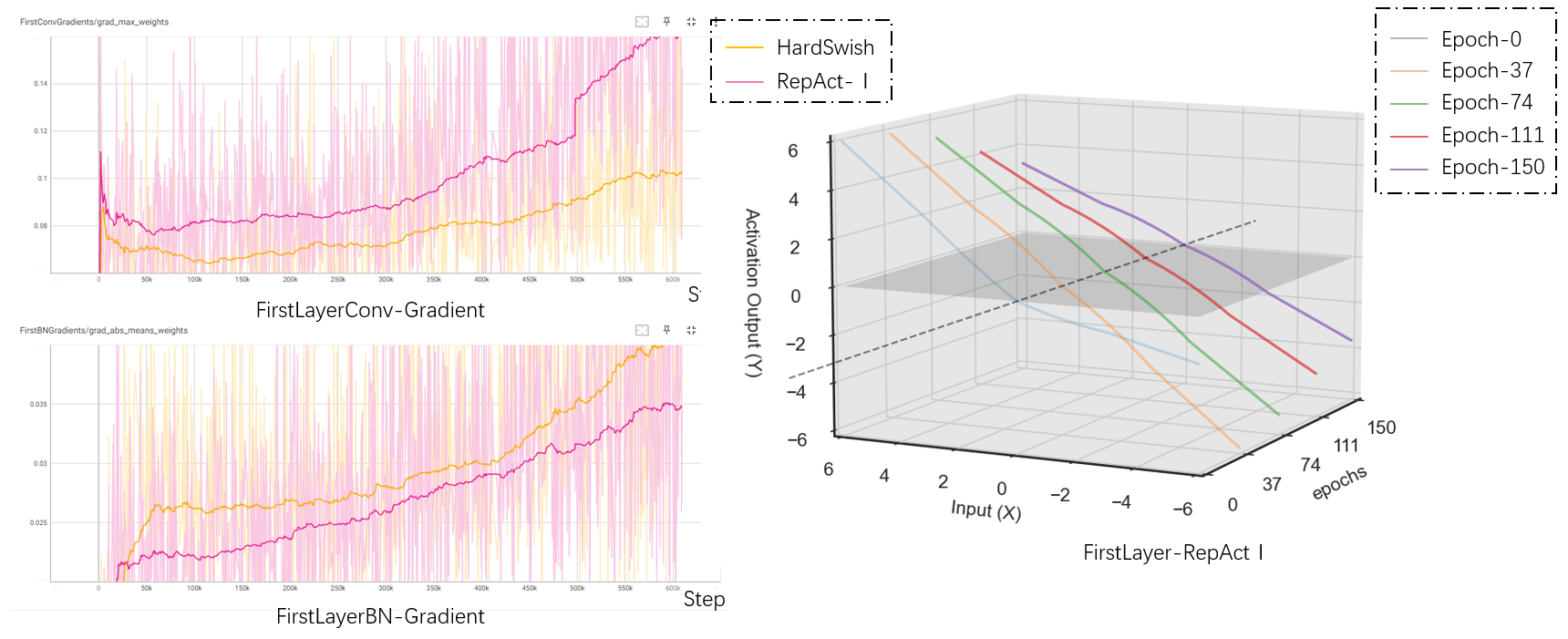}
        \caption{ Shallow gradient absolute mean training time curve and REPACT-I training visualization}
        \label{subfig:fig1}
    \end{subfigure}
    \vfill
    \begin{subfigure}[b]{\textwidth}
        \centering
        \includegraphics[width=0.8\textwidth]{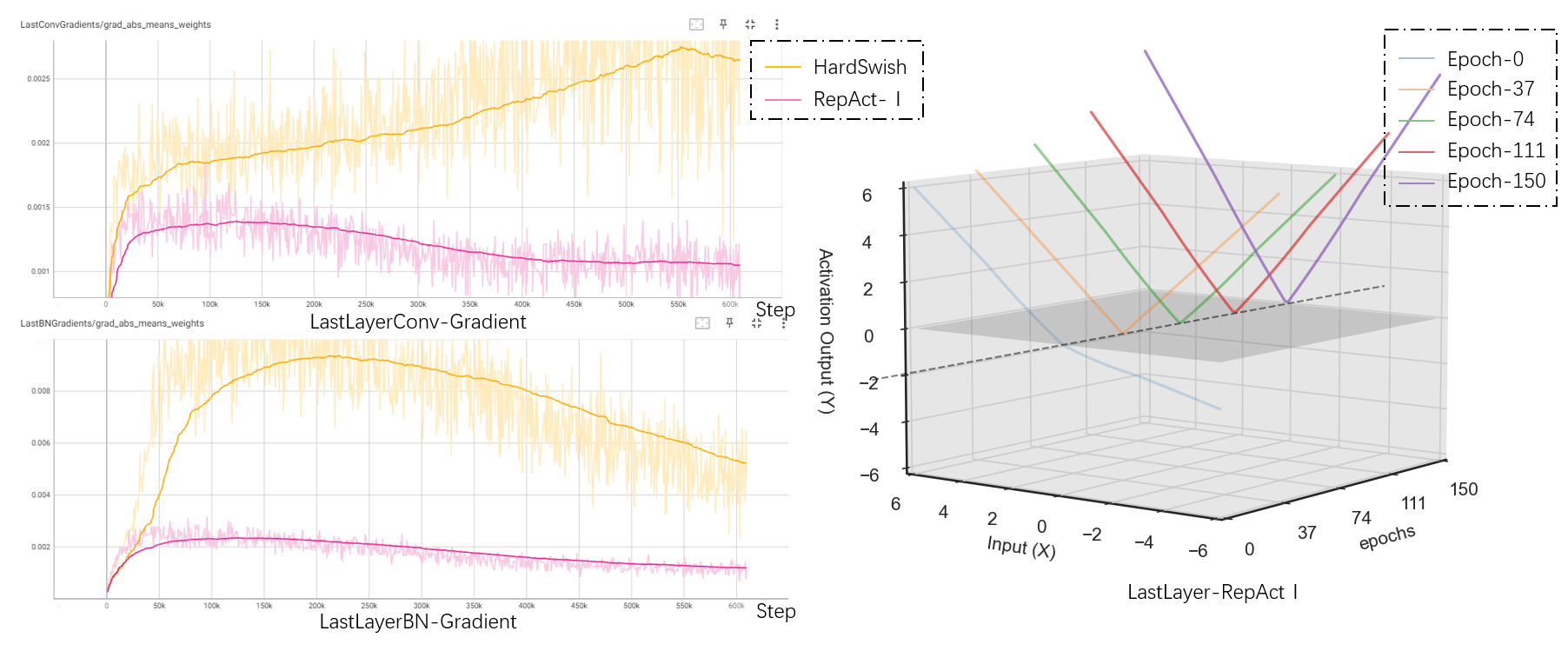}
        \caption{Deep gradient absolute mean training time curve and REPACT-I training visualization}
        \label{subfig:fig2}
    \end{subfigure}
    \caption{Figure.12 3D visualization of activation function selected by MobileNetV3-Large-RepAct I in ImageNet100 training with Epoch branch combination and corresponding layer Conv and BatchNorm gradient;}
    \label{fig:overall}
\end{figure}

We further observed the absolute average gradient curves in the first and last convoluted layers (Conv) and batch normalization (BN) within MobileNetV3-Large during training on ImageNet100, visualizing the adaptive effect of branch combination during training with RepAct I. As seen in Figure 12(a), the RepAct I branch combination in the shallow layers of the network tends towards linearity under the influence of gradients, preserving a rich texture information. In the deeper layers, as shown in (b), the last convolution layer's RepAct I adapts to a V-shaped activation function, akin to an absolute value function. This makes it easier for the gradients bearing classification labels (loss gradients) to propagate backwards to the middle and shallow layers of the network during backpropagation, which helps to alleviate the issue of class gradient vanishing, making the training of deep networks more stable. This corresponds to the visualized gradient curves in the deep layers' Conv, BatchNorm gradient training visualization, where the gradients of RepAct I (blue) are consistently higher than those of the original activation function (black).

\begin{figure}[ht!]
    \centering
    \begin{subfigure}[b]{0.45\textwidth} 
        \centering
        \includegraphics[width=\textwidth]{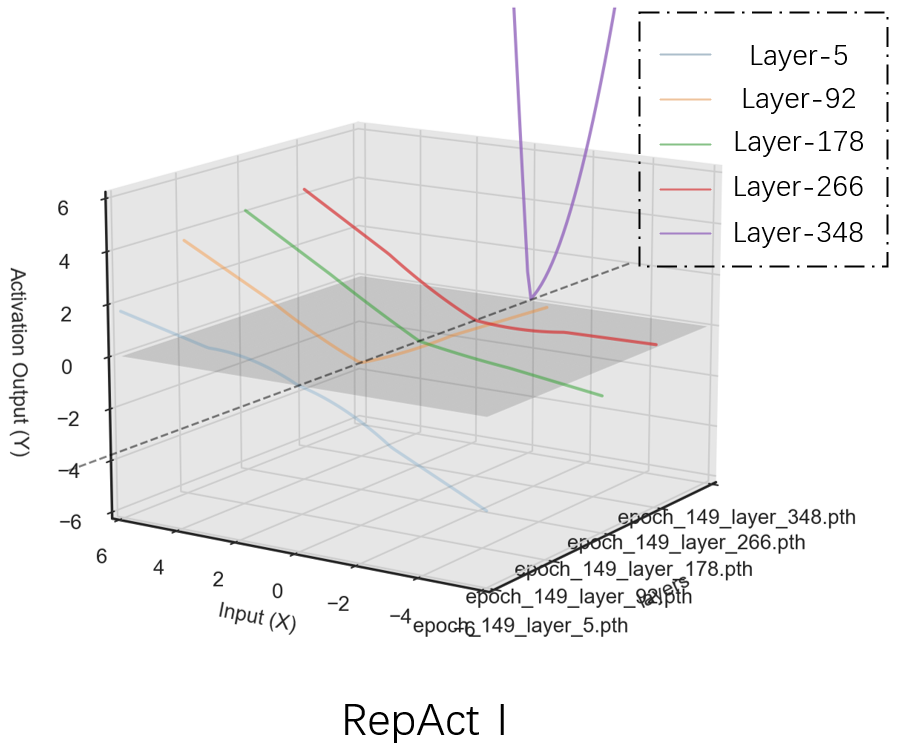}
        \caption{(a) shufflenet\textunderscore v2\textunderscore x0\textunderscore 5}
        \label{subfig:fig1}
    \end{subfigure}
    \hfill
    \begin{subfigure}[b]{0.45\textwidth} 
        \centering
        \includegraphics[width=\textwidth]{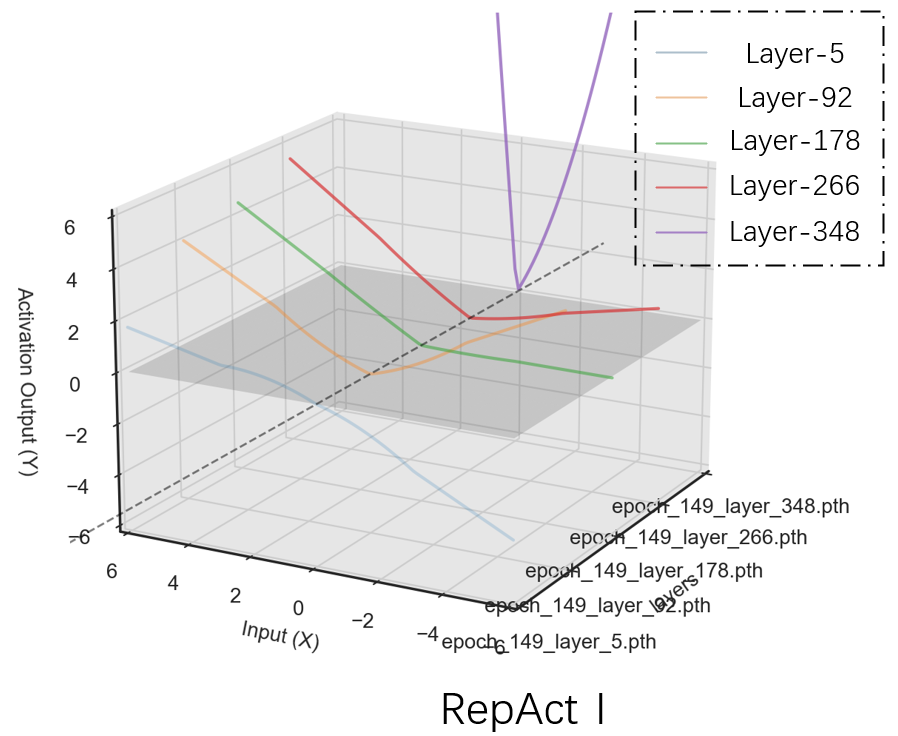}
        \caption{(b) shufflenet\textunderscore v2\textunderscore x1\textunderscore 0}
        \label{subfig:fig2}
    \end{subfigure}

    \vfill
    
    \begin{subfigure}[b]{0.45\textwidth} 
        \centering
        \includegraphics[width=\textwidth]{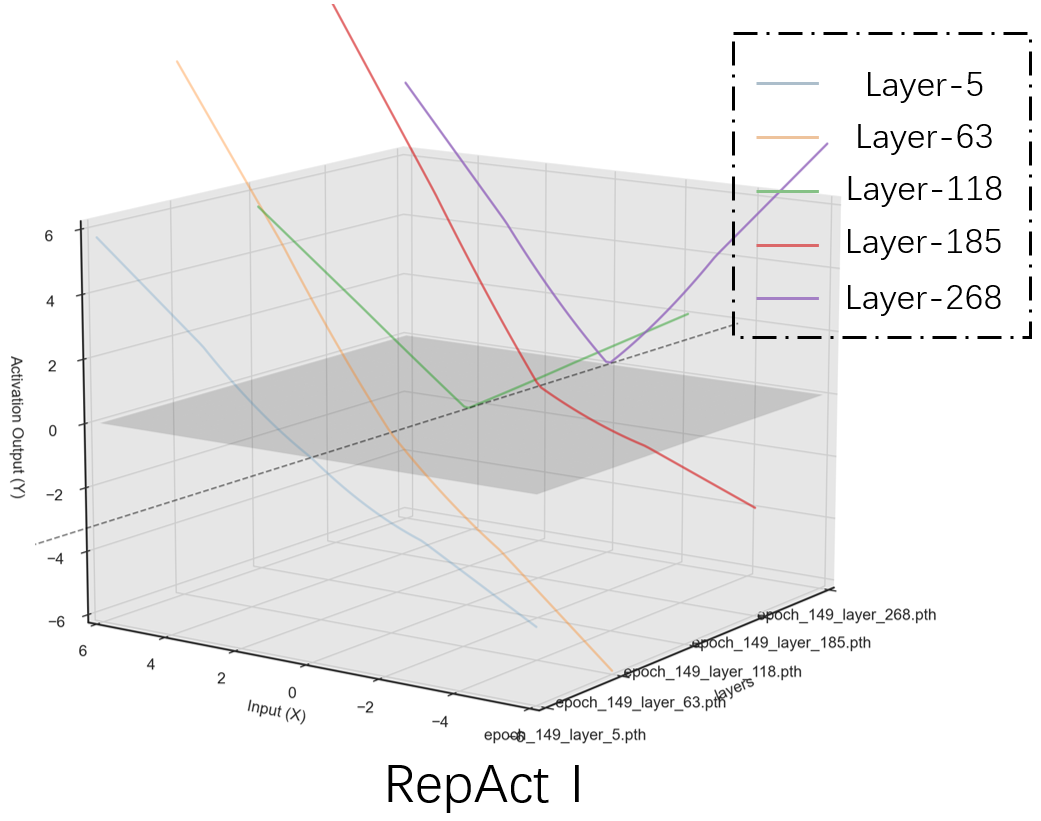}
        \caption{(c) MobileNetV3-Small}
        \label{subfig:fig3}
    \end{subfigure}
    \hfill
    \begin{subfigure}[b]{0.45\textwidth} 
        \centering
        \includegraphics[width=\textwidth]{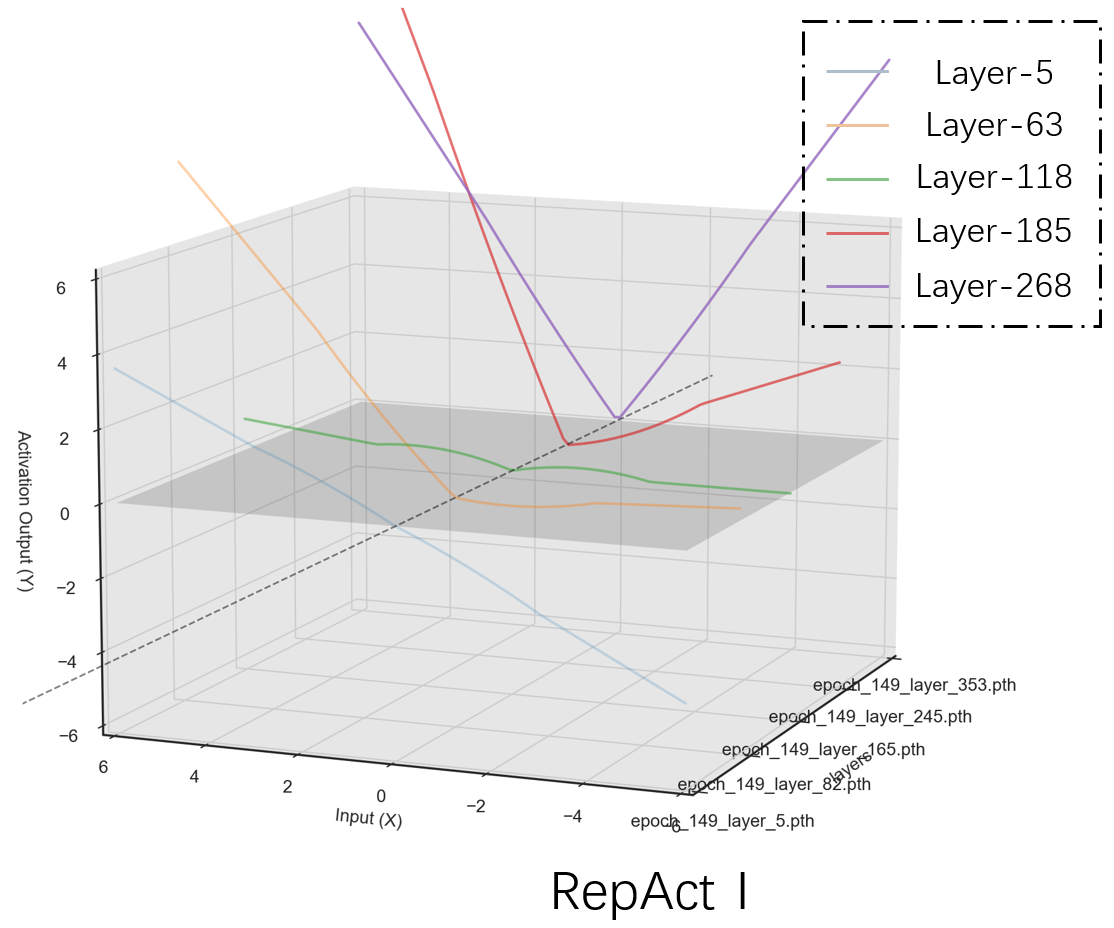}
        \caption{(d) MobileNetV3-Large}
        \label{subfig:fig4}
    \end{subfigure}

    \caption{Figure 13 RepAct I Visualization of different semantic layers of lightweight network after ImageNet100 training}
    \label{fig:overall}
\end{figure}

After the final training completion, in structurally similar networks such as MobileNetV3-Large and Small, as well as ShufflenetV2’s x0\textunderscore5 and x1\textunderscore0 versions, one can observe from Figure 13 the visualized results of five layers of RepAct I, ranging from the network's shallow texture layers to the deeper semantic layers. It is readily noticeable that in the first stem layer of both MobileNetV3 and ShufflenetV2 Large and Small,RepAct I adaptively leans towards linear branches, retaining a large amount of texture detail information in the shallow layers to be passed down to the deeper layers, which aligns with manual design experience. For the same type of network structure, the adaptive selection combination of RepAct I in similar semantic depth layers of the network is very consistent. This is even more pronounced in ShufflenetV2 changes only in parameter size. In the last layer, all adaptively form a function similar to the absolute value, ensuring that gradients carrying category information are effectively backpropagated.These findings indicate that RepAct adapts to choose and combine branches for differentiated activation in network layers of similar structure but different semantic levels. This adaptation ensures that the flow of inference information and the efficiency of gradient backpropagation are optimized, thus enhancing network learning capabilities and allowing lightweight networks to make fuller use of their inherent information capacity.

\subsection{RepAct commonality}

In other tasks where various lightweight networks act as the backbone, RepAct easily replaces the original activation function to enhance the backbone network's task learning capabilities and feature extraction abilities. Using the VOC12 dataset for object detection and segmentation \cite{50}, we divide the data following the original VOC12 division method for training and testing to demonstrate the universality of RepAct across different tasks.
\subsubsection{Target detection}

\begin{center}
\noindent Table 6: Comparison experiment mAP0.5 of each lightweight backbone of YOLOV5 on VOC12 dataset
\resizebox{1.0\linewidth}{!}{
\begin{tabular}{*{14}{c}}
  \toprule
  BackBone/Act  &Baseline&RepAct-I&RepAct-II&RepAct-III\\
  \midrule
    Shufflenetv2\_w0.5&	37.37(R)&	40.56\textcolor{red}{(+3.19)}&	39.64(+2.27)&	39.29(+1.92)\\
    Shufflenetv2\_w1	&48.37(R)&	49.82(+1.45)&	49.56(+1.19)&	49.86\textcolor{red}{(+1.49)}\\
    MNV3-large\_w0.5&	52.75(H)&	53.56\textcolor{red}{(+0.81)}&	53.51(+0.76)&	53.27(+0.52)\\
    MNV3-small\_w0.5&	45.08(H)&	46.25\textcolor{red}{(+1.17)}&	46.19(+1.11)&	46.20(+1.12)\\
  \bottomrule
\end{tabular}
}
\end{center}

We used YOLOV5 as the baseline framework, replacing the backbone part of the network with Shufflenetv2 and MobilenetV3, without additional adjustments to the hyperparameters. Detailed settings of the hyperparameters can be found in the appendix. We conducted comparative evaluations on the mAP0.5 metric between the original activation functions and RepAct, resulting in Table 6. For the original networks, the RepAct series showed a significant improvement on the mAP0.5 metric, and RepAct I was more suitable for these types of tasks.

\begin{center}
\noindent Table 7: Comparison experiment of mAP0.5 index of YOLOV5-MNV3-small\_w0.5 activation function on VOC12s
\resizebox{1.0\linewidth}{!}{
\begin{tabular}{*{14}{c}}
  \toprule
    Act&mAP0.5&Act&mAP0.5&Act&	mAP0.5&	Act&	mAP0.5\\
  \midrule
    HardSwish\cite{37}&	46.07&	ReLU\cite{1}&	-1.04&	Swish\cite{6}&	-0.59&	ELU\cite{23}&-2.13\\
    RepAct-I&	+0.3&	LReLU\cite{27}&	-1.75&	Mish\cite{24}&	-0.04&	SELU\cite{45}&	-5.70\\
    RepAct-II&	\textcolor{red}{+1.16}&	PReLU\cite{7}&	-0.41&	FReLU\cite{10}&	\textcolor{red}{+1.72}&	GELU\cite{46}&	-0.07\\
    RepAct-III&	\textcolor{red}{+0.96}&	AReLU\cite{8}&	-0.62&	DYReLU\cite{9}&	-0.47&	CELU\cite{47}&	-2.13\\
    Identity&	/&	Softplus\cite{48}&	-1.81&	ACONC\cite{31}&	+0.22&	SiLU\cite{46}&	+0.51\\
  \bottomrule
\end{tabular}
}
\end{center}

We conducted comparative experiments on the YOLOV5-MNV3-small\textunderscore5w0.5 model using various mainstream activation functions and their variants, as shown in Table 7. The RepAct series achieved accuracy in the object detection tasks comparable to other activation functions that require additional memory computations or exponential computational complexity, solely by using the computational complexity of the HardSwish form. Moreover, RepAct is more favorable for lightweight networks.
\subsubsection{Semantic segmentation}
\begin{center}
\noindent Table 8: Comparative experiments of DeepLabV3-MobileNetV3-Large on VOC12 data set
\resizebox{0.6\linewidth}{!}{
\begin{tabular}{*{14}{c}}
  \toprule
    Act.	&Mean\_IOU&	Global Acc\\
  \midrule
    BaseLine(HardSwish)\cite{37}&	37.9&	81.4\\
    RepAct-I&	40.0(+2.1)&	84.0(+2.6)\\
    RepAct-II&	40.1(+2.2)&	83.3(+1.9)\\
    RepAct-III&	42.5(+4.6)&	84.8(+3.4)\\
    ReLU\cite{1}&	36.8(-1.1)&	82.4(+1.0)\\
    LReLU\cite{27}&	35.2(-2.7)&	82.2(+0.8)\\
    PReLU\cite{7}&	40.7(+2.8)&	84.2(+2.8)\\
    FReLU\cite{10}&	29.4(-8.5)&	76.5(-4.9)\\
    DYReLU\cite{9}& 37.5(-0.4)&	82.8(+1.4)\\
    AReLU\cite{8}&	40.8(+2.9)&	84.3(+2.9)\\
    SiLU\cite{46}&	39.1(+1.2)&	83.5(+2.1)\\
    Softplus\cite{48}&36.0(-1.9)&78.1(-3.3)\\
    Swish\cite{6}&	39.6(+1.7)&	84(+2.6)\\
    Mish\cite{24}&	38.9(+1.0)&	83.1(+1.7)\\
    ELU\cite{23}&	38.0(+0.1)&	82(+0.6)\\
    SELU\cite{45}&	32.2(-5.7)&	79.3(-2.1)\\
    GELU\cite{46}&	38.0(+0.1)&	83.3(+1.9)\\
    CELU\cite{47}&	38.0(+0.1)&	82.3(+0.9)\\
    ACONC\cite{31}&	38.8(+0.9)&	83.6(+2.2)\\
  \bottomrule
\end{tabular}
}
\end{center}

In the task of semantic segmentation, we also replaced the backbone of DeepLabV3\cite{51} with MobileNetV3-Large as a lightweight backbone network (detailed hyperparameters settings can be found in the appendix). We compared the improvements of RepAct on the VOC12 segmentation dataset in terms of Mean\textunderscore IOU and Global CorRepAct indicators over the original activation functions. As shown in Table 8, thanks to the degradable global information of RepAct III, it performed better in semantic segmentation tasks, showing significant improvements compared to other activation functions.

\section{Summary and discussion}

This research introduces a series of plug-and-play, re-parameterizable, adaptive RepAct activation functions, taking full advantage of the characteristics of multi-branch training and single-branch inference. Without introducing extraneous parameters and additional memory computations, we have successfully enhanced the learning ability of lightweight networks by relying solely on power-function type RepAct. The mechanism behind RepAct was validated and analyzed from multiple perspectives including adaptive branch selection, forward feature scaling, and backward gradient propagation across different tasks and classical lightweight networks. However, there are still issues that necessitate further experimental exploration:

1.The selection of types of branches for RepAct, the balance of their quantities, and the degrees of re-parameterization still require further verification and experimentation. This paper has attempted re-parameterizable combinations of activation functions with power-function forms, and the exploration of combinations involving exponential-level activation functions remains to be explored.

2.For network structures with outstanding gradient design and learning capabilities, such as ResNet\cite{52}, using RepAct may lead to rapid convergence and overfitting. Therefore, how to balance the learning capabilities of RepAct or effectively regularize it remains a question that needs to be addressed.

\bibliographystyle{IEEEtran}

\appendix
\section{Appendix}
Experimental environment:
\begin{table}[h]
    \centering
    \begin{tabular}{| l | l |}
        \hline
        \textbf{Component} & \textbf{Configuration} \\
        \hline
        CPU & Intel Core i7-7700HQ \\
        \hline
        GPU & NVIDIA GeForce 3090Ti 24G \\
        \hline
        RAM & 16G \\
        \hline
        CUDA & 11.7 \\
        \hline
        Operating system & Ubuntu 22.04 x64 \\
        \hline
    \end{tabular}
    \caption{Hardware and Software Configuration}
    \label{tab:config}
\end{table}

Settings of training hyperparameters for each task dataset and partial network convergence curves:
ImageNet100 setup: image size is 224x224, epoch count is 150, batch size is 32, the learning rate starts at 0.01 with cosine annealing decay, CrossEntropyLabelSmooth is set to 0.1. Can be obtained from \href{https://github.com/2771096196/RepAct}{https://github.com/2771096196/RepAct}.

\begin{figure}[h]
    \centering
    \includegraphics[width=0.95\textwidth]{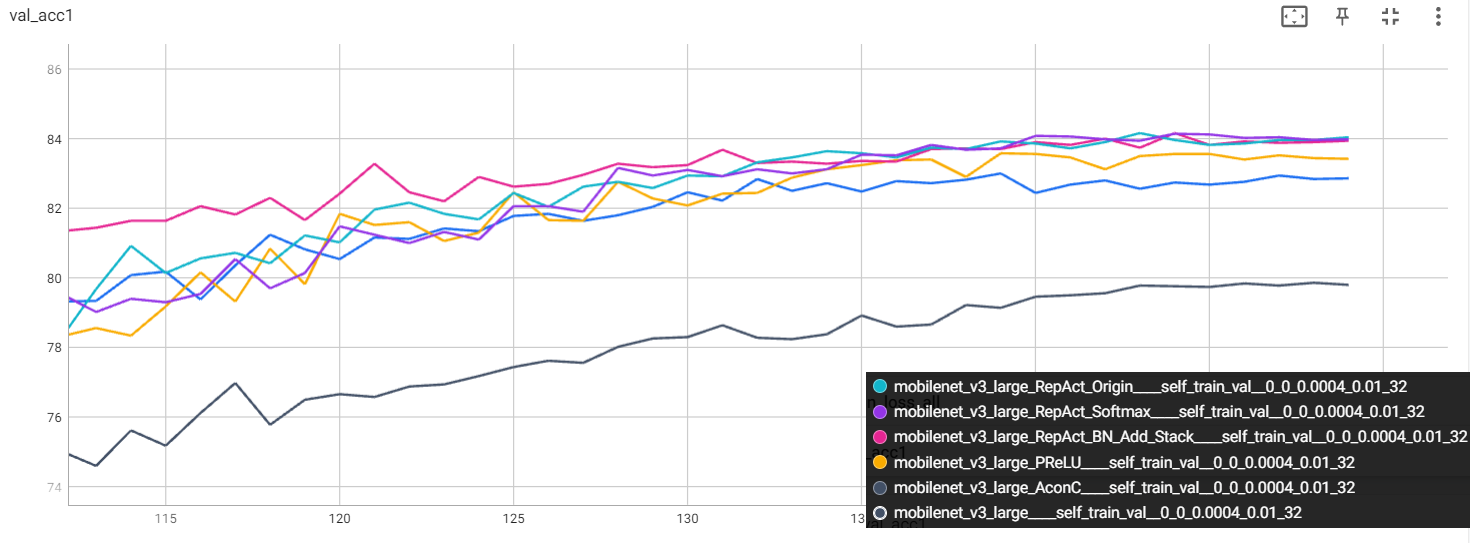}
    \caption{Convergence curve of activation functions for MobilenetV3Large series on ImageNet100}
    \label{fig:enter-label}
\end{figure}

Cifar100 setup: image size is 32x32, epoch count is 200, batch size is 128, the learning rate starts at 0.1 with step decay over epochs, includes a warm-up period of 1. For details, refer to the open-sourceproject at \href{https://github.com/weiaicunzai/pytorch-cifar100 }{https://github.com/weiaicunzai/pytorch-cifar100 }(reduce modifications to the downsampling rate of various lightweight networks to match the project).

\begin{figure}[h]
    \centering
    \includegraphics[width=0.95\textwidth]{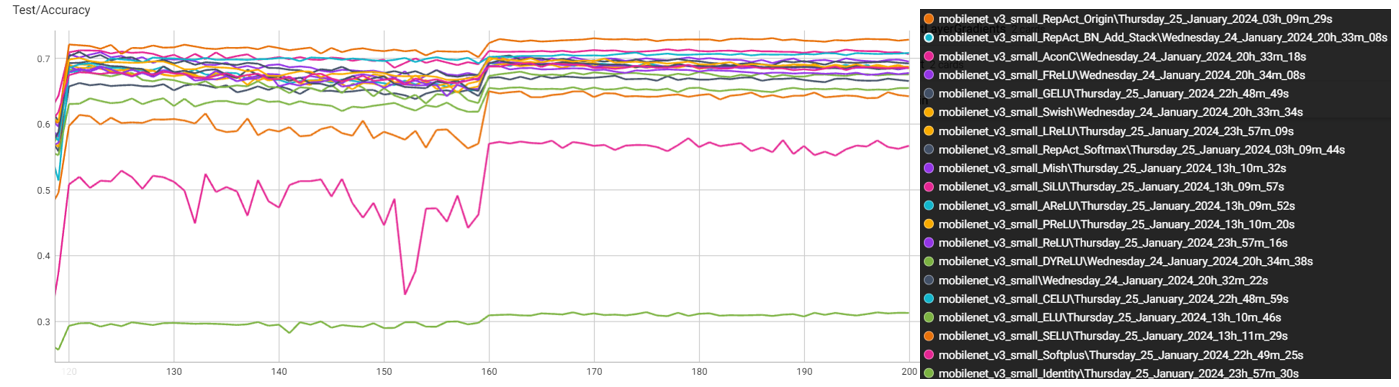}
    \caption{Figure 15: Convergence curve of activation functions for MobilenetV3Small series on Cifar100.}
    \label{fig:enter-label}
\end{figure}

\begin{figure}[h]
    \centering
    \includegraphics[width=0.95\textwidth]{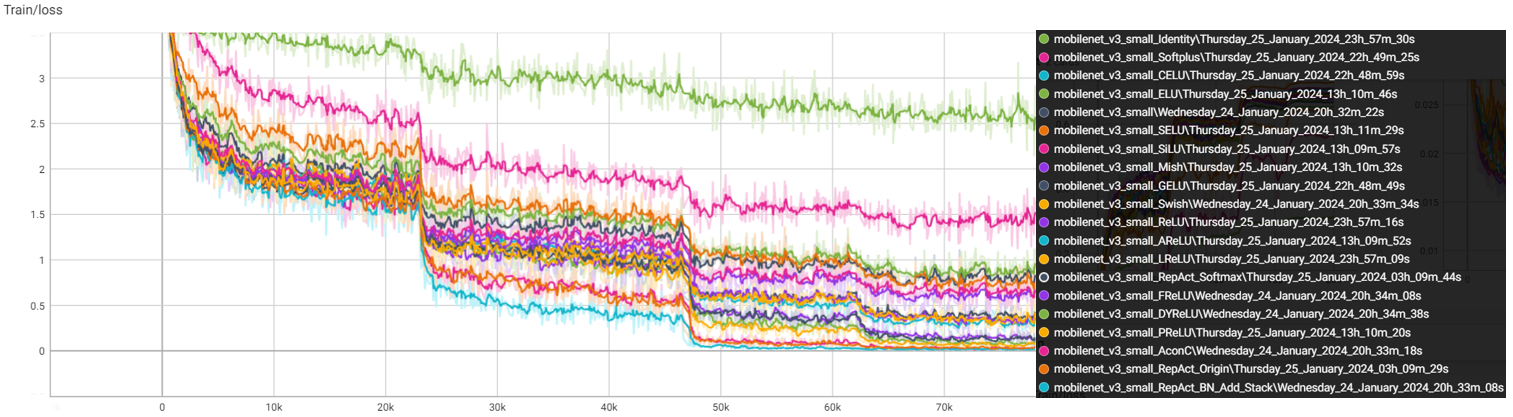}
    \caption{Figure 16: Convergence curve of TrainLoss for MobilenetV3Small activation functions on Cifar100.}
    \label{fig:enter-label}
\end{figure}

VOC12-DETECT setup: epoch count is 300, batch size is 8, image size is 640x640. For more details, refer to \href{https://github.com/ultralytics/yolov5.git}{https://github.com/ultralytics/yolov5.git}.

\begin{figure}[h]
    \centering
    \includegraphics[width=0.95\textwidth]{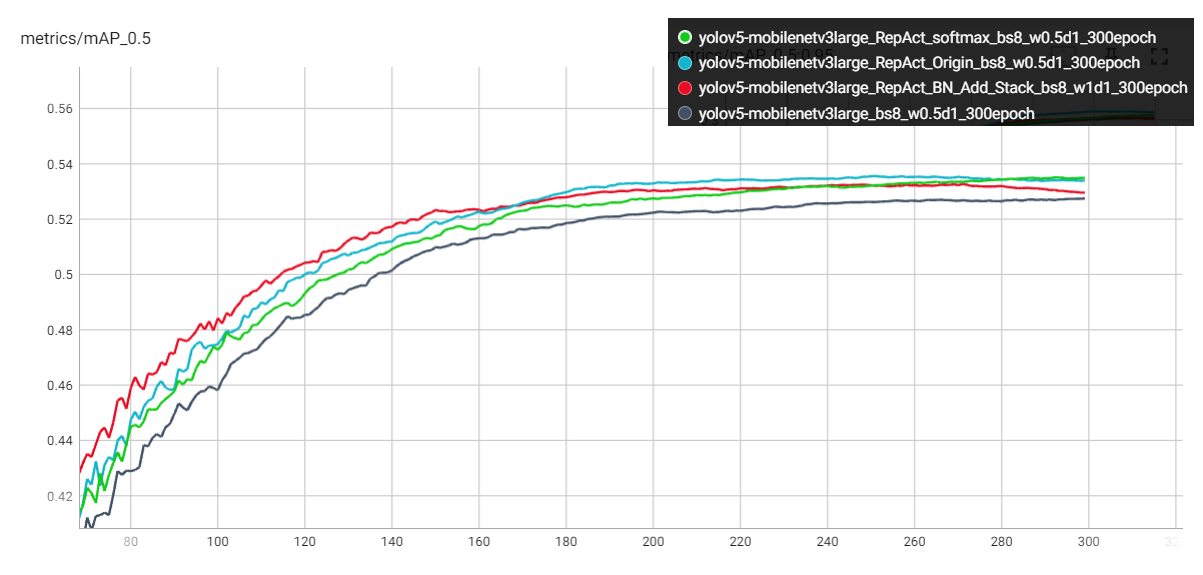}
    \caption{Figure 17: VOC12-DETECT-YOLOV5-Mobilenetv3Large series activation functions mAP\textunderscore50.5 convergence curve.}
    \label{fig:enter-label}
\end{figure}

VOC12-SEGMENT setup: epoch count is 300, batch size is 4, image size is 520x480. For further information, please refer to \href{https://github.com/WZMIAOMIAO/deep-learning-for-image-processing/tree/master/pytorch\textunderscore5v3segmentation/deeplab\textunderscore5v3}{https://github.com/WZMIAOMIAO/deep-learning-for-image-processing/tree/master/pytorch\textunderscore5v3segmentation/deeplab\textunderscore5v3}.

\begin{figure}[h]
    \centering
    \includegraphics[width=0.95\textwidth]{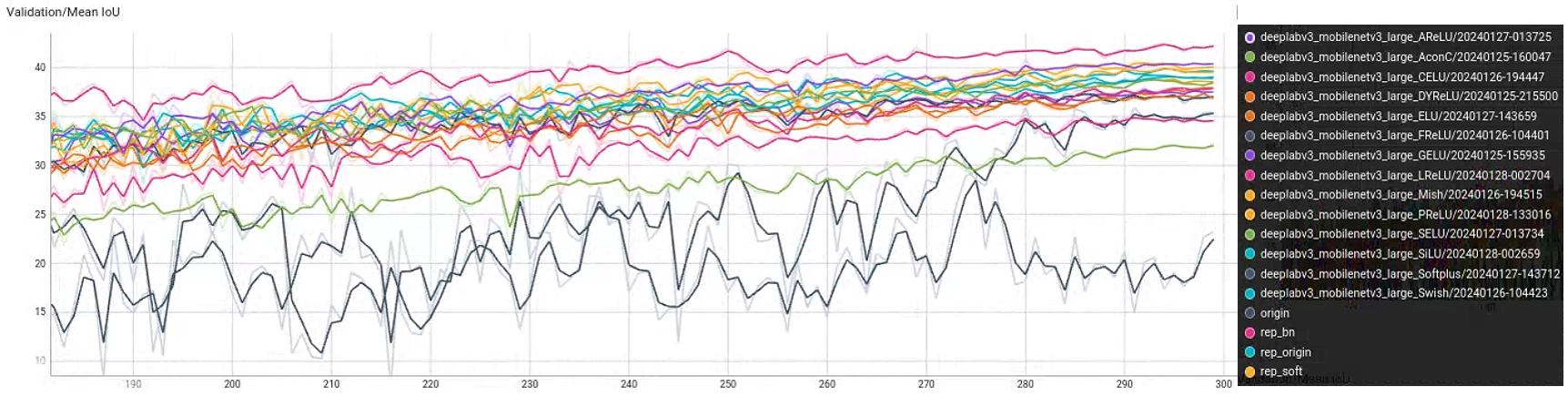}
    \caption{Figure 18: The convergence curve of various activation functions in VOC12-SEGEMENT-DeepLabV3-MobileNetV3-Large series.}
    \label{fig:enter-label}
\end{figure}
\end{sloppypar}
\end{document}